\documentclass[a4paper, 10pt, conference]{ieeeconf}      % Use this line for a4 paper

\IEEEoverridecommandlockouts% This command is only needed if 
\usepackage{times} % assumes new font selection scheme installed
\usepackage{amsmath} % assumes amsmath package installed
\usepackage{amssymb}  % assumes amsmath package installed
\usepackage{amsfonts}
\usepackage{siunitx}
\usepackage{algorithm}
\usepackage{algpseudocode}
\usepackage{graphicx}
\usepackage{float}
\usepackage[]{subfig}
\title{\LARGE \bf
    A Hierarchical Region-Based Approach for Efficient Multi-Robot Exploration
}

\author{{Di Meng\authorrefmark{2}\authorrefmark{3}, Tianhao Zhao\authorrefmark{1}\authorrefmark{3}, Chaoyu Xue\authorrefmark{2}, Jun Wu\authorrefmark{2} and Qiuguo Zhu\authorrefmark{1}\authorrefmark{2}\authorrefmark{4}}
    \thanks{\authorrefmark{2} Di Meng, Chaoyu Xue, Jun Wu and Qiuguo Zhu are with the College of Control Science and Engineering, Zhejiang University, Hangzhou, China, email: \texttt{\{monty, xuechaoyu, junwuapc\}@zju.edu.cn}.}%
    \thanks{\authorrefmark{1} Tianhao Zhao and Qiuguo Zhu are with the Polytechnic Institute, Zhejiang University, Hangzhou, China, email: \texttt{\{zhaothianhao, qgzhu\}@zju.edu.cn}.}%
    \thanks{\authorrefmark{3}These authors have equally contributed to the work.}%
    \thanks{\authorrefmark{4} Corresponding author.}%
    \thanks{This work was supported by the National Key R\&D Program of China (Grant No. 2022YFB4701502), the ``Leading Goose'' R\&D Program of Zhejiang (Grant No. 2023C01177), the Key R\&D Project on Agriculture and Social Development in Hangzhou City (Asian Games) (Grant No. 20230701A05), and the Key Research Project of Zhejiang Lab (Grant No. 2021NB0AL03).}
}

\begin{document}
\maketitle
\thispagestyle{empty}
\pagestyle{empty}

\begin{abstract}
    Multi-robot autonomous exploration in an unknown environment is an important application in robotics.
    Traditional exploration methods only use information around frontier points or viewpoints, ignoring spatial information of unknown areas.
    Moreover, finding the exact optimal solution for multi-robot task allocation is NP-hard, resulting in significant computational time consumption.
    To address these issues, we present a hierarchical multi-robot exploration framework using a new modeling method called RegionGraph.
    The proposed approach makes two main contributions:
    1) A new modeling method for unexplored areas that preserves their spatial information across the entire space in a weighted graph called RegionGraph.
    2) A hierarchical multi-robot exploration framework that decomposes the global exploration task into smaller subtasks, reducing the frequency of global planning and enabling asynchronous exploration.
    The proposed method is validated through both simulation and real-world experiments, demonstrating a 20\% improvement in efficiency compared to existing methods.
\end{abstract}

\section{INTRODUCTION}\label{sec:introduction}

Autonomous exploration is a widely researched challenge in robotics, with practical applications in various real-world scenarios.
Multi-robot exploration is an area of considerable interest due to its efficiency benefits over single robots.

Although progress has been made, there are still several unresolved issues that require further investigation.
Two of these issues are of particular concern to us.
Firstly, traditional exploration approaches are often based on frontier points or viewpoints~\cite{yamauchi_frontier-based_1997,gao2018improved}, which only consider information around a limited number of points.
These approaches overlook the information of unexplored areas at a global level and lead to a local optimum.
Secondly, obtaining the exact optimal solution for multi-robot task allocation is NP-Hard~\cite{khamis2015multi}. Solutions between two neighboring decisions can become unstable, leading to robot backtracking. This greatly increases exploration time and reduces efficiency.

To address these issues,
this work introduces a novel hierarchical multi-robot exploration framework that overcomes limitations in traditional approaches by globally incorporating spatial information of unexplored areas.
By integrating a RegionGraph, our method enables efficient task allocation while preserving spatial charateristics, reducing computational overhead and improving exploration efficiency. 
The statistical results demonstrate that our method outperform the state-of-the-art by more than 20\% across multiple metrics

The proposed hierarchical framework consists of one center node and several client nodes.
The center node divides the entire exploration task into sub-tasks for client nodes to execute.
The main contributions of this paper lie in:
\begin{itemize}
    \item A novel modeling method is introduced to represent unexplored areas.
          This method generates a weighted graph, named RegionGraph, to store data on unexplored areas throughout the entire space. This enables subsequent task allocation to consider more information, thus facilitating modeling and improving solution quality.

    \item A novel hierarchical exploration framework that decomposes the global exploration task into smaller subtasks.
          The framework reduces the impact of unstable solutions by decreasing the frequency of global decision making and minimizes unnecessary waiting time through asynchronous exploration. This results in more efficient exploration.
\end{itemize}
\begin{figure}[t]
    \centering
    \includegraphics[clip,width=\columnwidth]{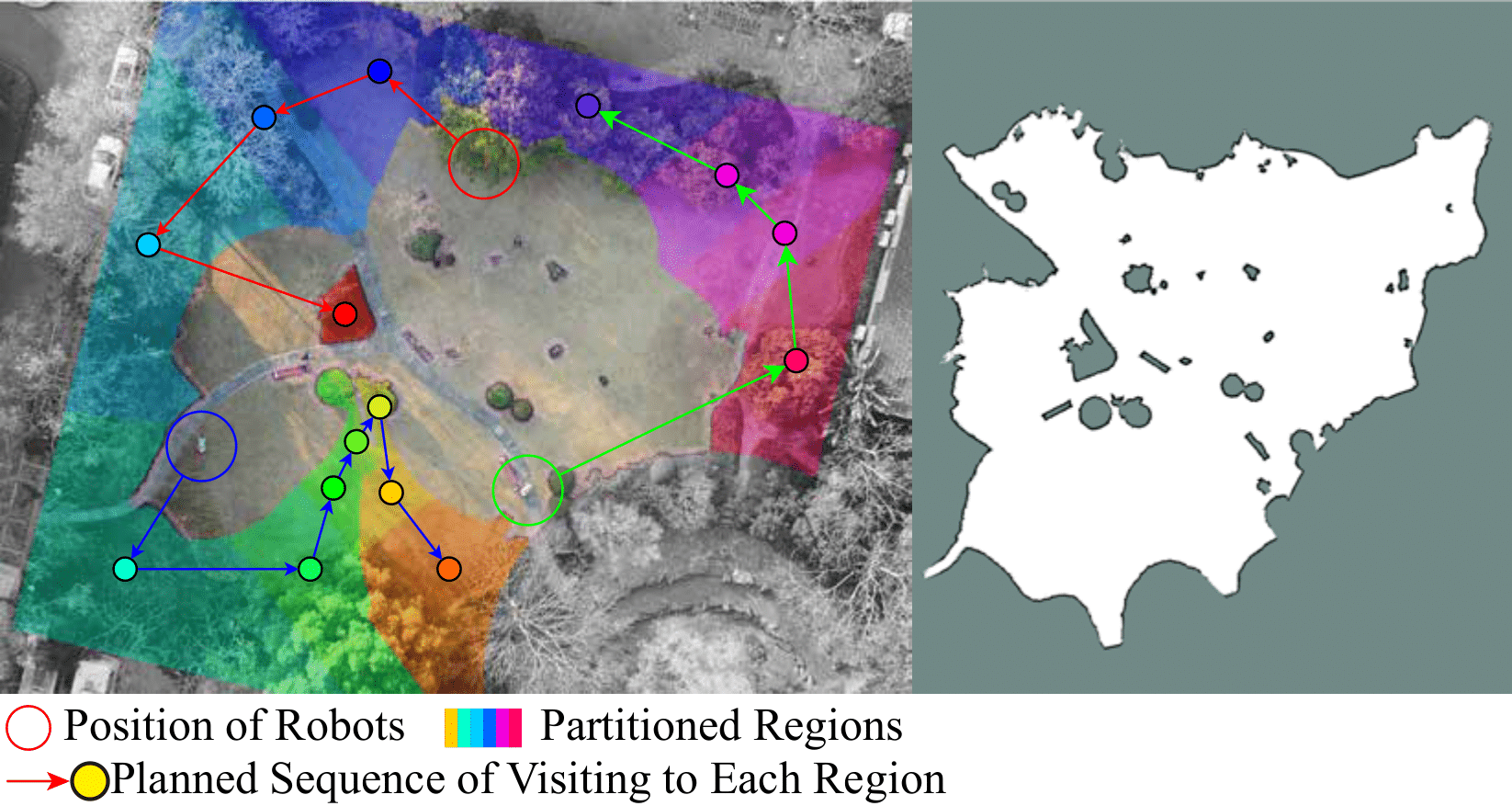}
    \caption{Results of a task allocation (left) and the explored map (right) in the real-world experiment.
        The image on the left displays a planning result on a global scale. The colored areas represent partitioned regions. The gray area represents uninterested areas outside the task.}\label{fig:frontpage}
    \vspace{-1em}
\end{figure}

This paper will be presented in the following manner:
Section~\ref{sec:related_works} summarizes previous related works.
Section~\ref{sec:problem_definition} presents the problem definition and some hypothetical conditions.
Section~\ref{sec:system_overview} introduces the overall structure of the proposed method, followed by Section~\ref{sec:methodology} which describes two main contents of the approach.
Simulation and real-world experiments are displayed in Section~\ref{sec:simulation_and_experiments}, and finally Section~\ref{sec:conclusion} concludes this paper and discusses potential improvements in future work.

\section{RELATED WORKS}\label{sec:related_works}

The process of multi-robot exploration is carried out in an iterative manner.
In each cycle, robots receive new environmental information from sensors.
Two crucial problems must be addressed: identifying next targets for exploration and determining which robot should be assigned to each target.
This involves two phases: target generation and target allocation.

\subsection{Target Generation}
Since Yamauchi introduced the concept of frontier~\cite{yamauchi_frontier-based_1997}, selecting frontier points as targets has become a common method for target generation.
Some improvements to this approach have been proposed.
Gao~\cite{gao2018improved} improved the method by incorporating angle information into frontier points.
Abony~\cite{abonyi2023autonomous} transformed frontier points into a tree structure, which provides a path for robot backtracking.

Sample-based method is an alternative approach to target generation.
Some of this kind are based on Rapidly-exploring Random Trees (RRT), which is a tree structure that extends from known to unknown areas.
One notable example is the RH-NBV method proposed by Bircher~\cite{bircher_receding_2016}.
Umari~\cite{umari_autonomous_2017} implemented a real-time local RRT and incrementally updated a global RRT for multi-robot exploration.
Dang~\cite{dang_graph-based_2020} constructed a Rapidly Random Graph (RRG) along the vehicle trajectory.

Some methods combined two or more techniques to generate target points more effectively.
For instance,~\cite{dai2020fast} combined a frontier-based approach with a sampling-based method to select target points from the safe area surrounding robots.
Zhou~\cite{zhou_fuel_2021} sampled directly near detected frontiers to obtain the best viewpoint.
Zhang~\cite{zhang2021multi} combined the RRT with an Artificial Potential Field (APF) method.
They utilized the APF method to guide the selection of sampling points based on the RRT method.

\subsection{Target Allocation}\label{subsec:related_works_object_allocation}
The target points generated initially are often numerous. It is necessary to cluster them into frontier clusters~\cite{zhou_fuel_2021,solanas2004coordinated} or regions~\cite{bi2023cure}.
There are two methods that can be used for task allocation: the greedy method and the Traveling Salesman Problem (TSP) method.
The greedy method involves sorting targets based on a cost or utility function and selecting the highest ranked one while discarding the rest~\cite{umari_autonomous_2017,mannucci_autonomous_2018,burgard_coordinated_2005}.
The utility function usually requires balancing between distance and information gain, as demonstrated in~\cite{yamauchi1998frontier,burgard2000collaborative}. Other approaches take into account additional information.
Zhou~\cite{zhou_fuel_2021} incorporated the turning angle between adjacent targets into cost function, while Pimentel~\cite{pimentel2018information} selected the most information-rich frontier from a probabilistic perspective. The TSP based method consider all targets in a decision-making round, determining a visiting sequence of all targets. It factors in longer-term considerations, as employed in~\cite{zheng_hierarchical_2022} and~\cite{ibrahim2021enhanced}.

For the multi-robot exploration, integrated methods have been proposed.
Dong~\cite{dong_multi-robot_2019} modeled the problem as an Optimal Massive Transport (OMT) problem, solving the OMT problem by a K-means clustering method.
Bi~\cite{bi2023cure} used a Voronoi partitioning method based on the position of robots to maintain local exploration windows.
Sun~\cite{sun_hierarchical_2023} employed a hierarchical approach, dividing exploration process into two distinct phases.
Some learning based methods have also been proposed~\cite{kamalova2022occupancy, alitappeh2022multi}.

\section{PROBLEM DEFINITION}\label{sec:problem_definition}

Let $\Omega$ represent the entire unknown environment aimed to explore. The exploration task focuses on mapping the observable environment $\Omega_o$, which is defined as $\Omega_o = \Omega \setminus \Omega_u$, where $\Omega_u$ denotes unobservable areas (e.g., rooms without entrances) that typically constitute a small portion of $\Omega$ due to robot capability limitations.

Our method requires a predefined Region of Interest (ROI) to determine the exploration range, without compromising the assumption of no prior knowledge about the environment's specific layout. The spatial information is r epresented as a two-dimensional occupancy grid map $M$, consisting of free cells $M_f$, obstacle cells $M_o$, and unknown cells $M_u$.

The robot team is denoted as $\mathcal{R} = \{R_1, R_2, \ldots, R_N\}$. Each robot $R_i$ is equipped with sensors (e.g., a radar with effective radius $r$ and an odometer) that gather environmental data. Through a Simultaneous Localization and Mapping (SLAM) module, each robot provides its local map $M_i(t)$ and position $p_i(t) = (x_i(t), y_i(t), \theta_i(t))$ at time $t$.

The proposed method focus on designing an efficient exploration strategy to enhance coordination and exploration efficiency. When any robot $R_i$ requires a target, this strategy takes as input the current known information $\{\{M_i(\tau)\}, \{p_i(\tau)\} | i = 1,\ldots,N; \tau = 0,\ldots,t\}$ and calculates target point  $\hat{p}_i = (x_i, y_i)$ for robot to explore unknown areas within the ROI. A navigation module guides robots to these target points.

It is assumed that perfect inter-robot communication and that local collision avoidance is handled by lower-level navigation systems, allowing us to focus on core exploration and mapping strategies.

The problem is thus defined: Given an unknown environment $\Omega$ and a robot team $\mathcal{R}$, develop an exploration strategy that generates a sequence of target points $\{\hat{p}_i\}$ to guide the robots in mapping $\Omega_o$. The goal is to progressively reduce unknown areas $M_u$ until $\Omega_o$ is entirely mapped.
\begin{figure*}[t]
    \centering
    \includegraphics[width=\linewidth]{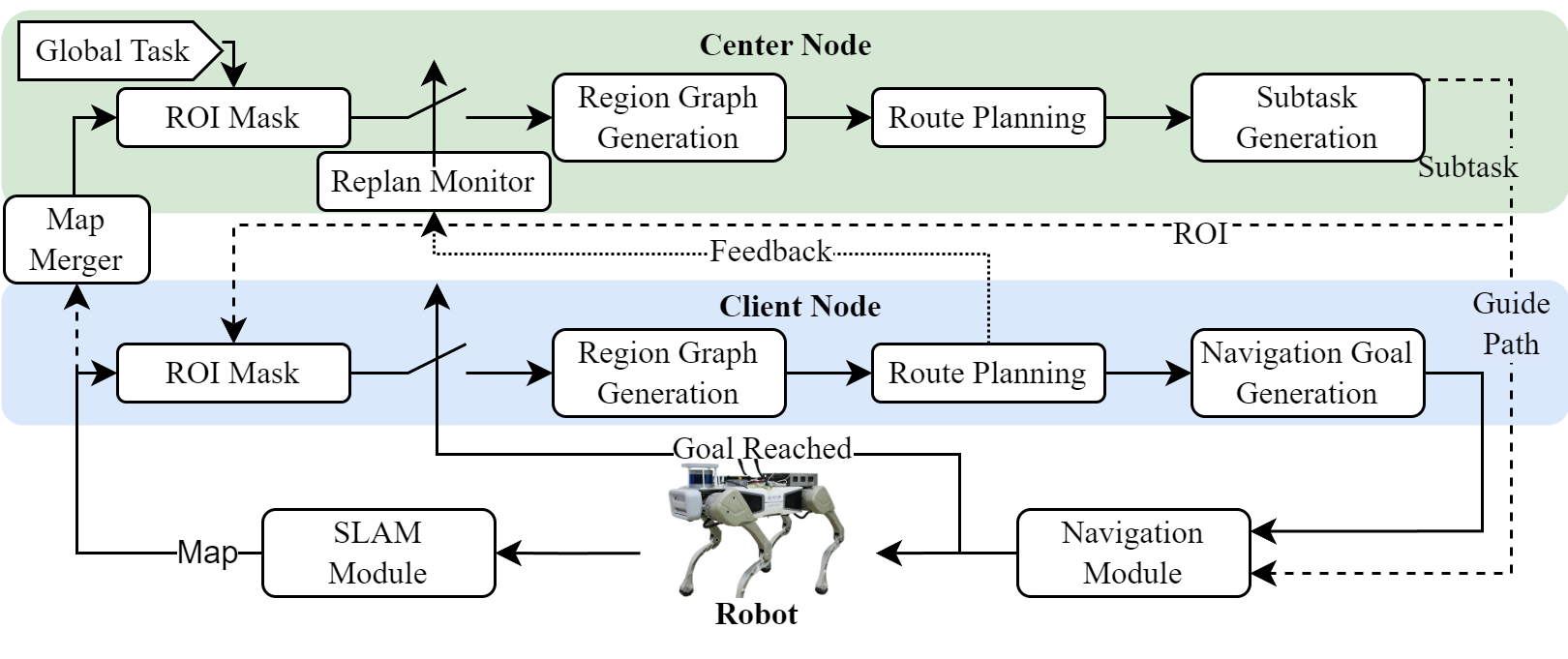}
    \vspace{-2em}
    \caption{System Overview}\label{fig:SystemOverview}
    \vspace{-1em}
\end{figure*}
\section{SYSTEM OVERVIEW}\label{sec:system_overview}

The proposed method, illustrated in Fig.~\ref{fig:SystemOverview}, comprises an exploration algorithm using RegionGraph and a hierarchical framework for independent agent exploration. 
%The exploration algorithm first generates a RegionGraph by extracting frontiers from the occupancy grid map, partitioning the unexplored area into regions, and constructing a graph representing these regions and their connections. It then plans routes by modeling task allocation as a Vehicle Routing Problem (or Traveling Salesman Problem for single robots) to determine the optimal region visitation order.
The proposed algorithm generates a RegionGraph by extracting frontiers from the occupancy grid map, partitioning unexplored areas into regions and representing their spatial connections. This enables task allocation modeled as a Vehicle Routing Problem , optimizing region visitation while preserving spatial information.

The hierarchical framework comprises a center node for global task allocation and multiple client nodes for local task refinement. The center node processes comprehensive global information and generates subtasks based on route planning result. Meanwhile, client nodes further decompose these subtasks, performing local task allocation to generate specific navigation targets. These client nodes continuously report their progress to the center. Notably, client nodes operate asynchronously, continuing exploration during the center node's replanning, thus enhancing system efficiency and robustness.
\section{METHODOLOGY}\label{sec:methodology}
\begin{figure*}[thbp]
    \vspace{0.4em}
    \centering
    \includegraphics[width=\linewidth]{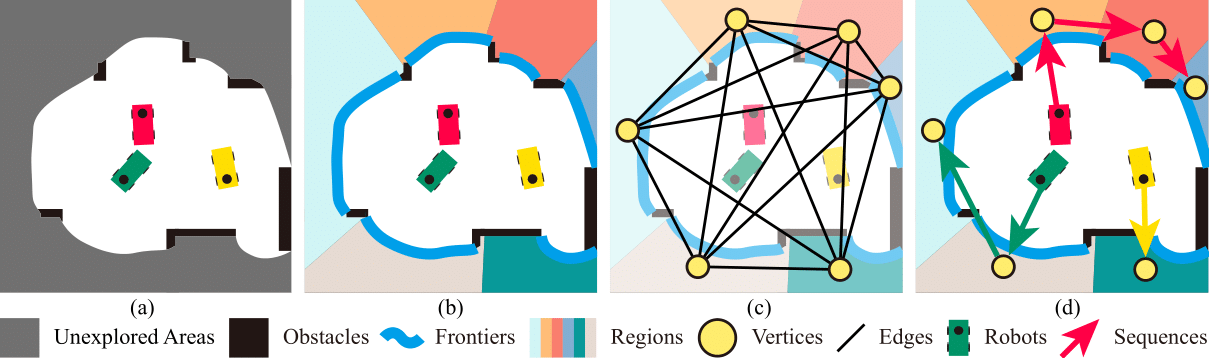}
    \caption{The Exploration Algorithm Using the RegionGraph. (a) Map updates received. (b) Unexplored areas are first partitioned using a Voronoi-like method. (c) Then, a RegionGraph is constructed, with each vertex representing a region and each edge representing a connection. (d) Finally, the optimal visiting sequences can be obtained after solving the VRP.}\label{fig:exploration_overview}
    \vspace{-1em}
\end{figure*}
\subsection{Exploration Algorithm Using RegionGraph}\label{sec:RegionGraph}
An algorithm for exploration is developed using the RegionGraph, as shown in Fig.~\ref{fig:exploration_overview}. It begins by partitioning the regions necessary to construct the RegionGraph. Next, the RegionGraph is constructed, followed by VRP modeling and route planning using the RegionGraph. Finally, the optimal visiting sequences for regions and a guide path is generated.

\subsubsection{Regions Partition}

\begin{algorithm}[t]
    \caption{Region Partitioning}\label{alg:region_graph_construction}
    \begin{algorithmic}[1]
        \Require Map $M$, Rapidly-exploring Random Graph $RRG$,
        \Require Maximum iteration limit  $N \in \mathbb{N}^+$, sampling radius $r \in \mathbb{R}^+$, field of view threshold $t_{fov} \in [0, 1]$
        \Ensure RegionGraph $G$
        \State $F \gets \textsc{DetectFrontiers}(M)$
        \State $P_v \gets \emptyset$ \Comment{Initialize viewpoints set}
        \ForAll{$f \in F$}
        \State $P_{cv} \gets \emptyset$ \Comment{Initialize candidate viewpoints set}
        \For{$i \gets 1$ \textbf{to} $N$}
        \State $p \gets \textsc{RandomSampling}(f, M, r)$
        \If{$\textsc{IsAccessible}(p, M)$}
        \If{$\textsc{FOV}(p, f) > t_{fov}$}
        \If{$\textsc{CanAddToRRG}(p, RRG)$}
        \State $RRG.\textsc{AddVertex}(p)$
        \State $P_{cv} \gets P_{cv} \cup \{p\}$
        \EndIf
        \EndIf
        \EndIf
        \EndFor
        \If{$P_{cv} \neq \emptyset$}
        \State $P_v \gets P_v \cup \{\textsc{MaxFOVPoint}(P_{cv})\}$
        \EndIf
        \EndFor
        \State $\textit{Regions} \gets \textsc{VoronoiPartition}(P_v, M)$
        \State $G \gets \textsc{ConstructRegionGraph}(\textit{Regions}, RRG)$
        \State \Return $G$
    \end{algorithmic}
\end{algorithm}

The generation of a RegionGraph involves partitioning unexplored areas into distinct regions, as outlined in Algorithm~\ref{alg:region_graph_construction}. This process begins with frontier detection upon receiving a map update, which extracts contours of unexplored areas, subtracts obstacles, and clusters frontier points based on a predefined sensor radius. A Rapidly-exploring Random Graph (RRG) maintains a dynamic topological representation of traversable areas, expanding as the map is updated.

For viewpoint determination, the algorithm employs uniform random sampling near frontiers. These candidate points undergo accessibility and field-of-view (FOV) checks, with those failing discarded. Remaining points are added to the RRG in descending FOV order, with the first successfully added point chosen as the frontier's viewpoint. Frontiers unable to generate valid viewpoints are eliminated.The surviving frontiers serve as seeds for dividing unexplored areas using a Voronoi-like method. Each region $s_i$ is defined as:
\begin{equation}
    s_i = \{ c \in M_\text{u} \mid d(c,f_i) \leq d(c,f_j), \forall f_j \neq f_i \in F \}
\end{equation}
where $M_\text{u}$ is the unexplored map area, $c$ is an unexplored cell, $F$ is the frontier set, and $d(c,f) = \min\limits_{c_i \in f} ||c-c_i||_2$ represents the minimum Euclidean distance from cell $c$ to frontier $f$.
\subsubsection{RegionGraph}

The RegionGraph $G$ is formally defined as a weighted graph:
\begin{equation}
    G = (V, E, w, d)
\end{equation}
where $V$ is the set of vertices, each representing a region. $E \subseteq V \times V$ is the set of edges, representing spatial connections between regions. $w: V \rightarrow \mathbb{R}^+$ is a vertex weight function, estimating the exploration workload of each region. $d: E \rightarrow \mathbb{R}^+_0$ is an edge weight function, representing the travel cost between regions.

Each vertex $v \in V$ is a tuple $v = (s, f, p_v)$, $s$ is the region, $f$ is the frontier of the region, and $p_v$ is the viewpoint (entry point) of the region. The set of vertices and edges are defined as:
\begin{align}
    V & = \{(s, f, p_v) \mid s \in S\}                 \\
    E & = \{(v_i, v_j) \mid v_i, v_j \in V, i \neq j\}
\end{align}
where $S$ is the set of all regions.

The vertex weight function $w$ estimates the exploration workload for each region. For a vertex $v_i$, the weight $w(v_i)$ is calculated as:
\begin{equation}
    w(v_i) = \frac{\sum_{j=1}^{N} d_j}{A_\text{explored}} \cdot A_i
\end{equation}
where $N$ is the number of robots. $d_j$ is the total distance traveled by robot. $j$ since the start of exploration. $A_\text{explored}$ is the total area explored. $A_i$ is the area of the region represented by vertex $v_i$.
This equation estimates the workload by first calculating the average distance required to explore a unit area, then multiplying it by the area of the unexplored region.

The edge weight function $d$ represents the travel cost between regions. For an edge $(v_i, v_j) \in E$, $d(v_i, v_j)$ is the distance between the entry points of the regions represented by $v_i$ and $v_j$. Note that $d(v_i, v_j) = 0$ if a robot can directly move from one region to the other without additional travel.
\subsubsection{VRP Modeling and Route Planning}\label{Workload_Estimation}
The problem of task allocation can be modeled as a VRP using the RegionGraph after two adjustments.
Firstly, a cost matrix needs to be constructed for solving VRP. The weight of the vertex in the RegionGraph is redistributed to the connecting edges using the transformation shown in (6).
\begin{equation}
    d^\star_{ij} = \frac12 \left( \hat{w}_i + \hat{w}_j \right) + d_{ij}
\end{equation}
Equation (6) updates the weight of edges $(v_i,v_j)$ to $d^\star_{ij}$, while the weights of the connecting vertices are represented by $\hat{w}_i$ and $\hat{w}_j$, and the original weight of edge is $d_{ij}$. This adjustment ensures the distance is maintained the same as robots enter and exit a region.
Secondly, robots are introduced to the cost matrix. To achieve non-return VRP, the weight of the robot to each region is set to the distance it needs to travel to reach the region. The weight from the region to the robot and the weight between robots are set to zero. This modification results in a fully defined VRP, which is ready for being calculated by a VRP solver~\cite{perron_or-tools_2023}.

The solution to the VRP problem is an optimal visiting sequence for the region of vertices in the RegionGraph. The proposed method generates a guide path in the RRG for each non-empty sequence using Johnson's shortest path algorithm, leading to the first goal in the sequence.

\subsection{Hierarchical Exploration Framework}

The exploration framework consists of a two-tier system comprising a center node and multiple client nodes.
The center node is responsible only for global decision making, generating subtasks rather than specific target points.
Each client node is deployed on a robot entity to make decisions, sending goals to the navigation module of the robot.
They are responsible for performing subtasks dispatched from the center node. The client nodes are completely asynchronous to the center node.
When the center node is replanning, a client node can continue executing the previous subtask until it receives a new one, without having to stop and wait.

\subsubsection{Task Hierarchical Mechanism}
\begin{figure}[t]
    \vspace{-0.9em}
    \centering
    \subfloat[]{\includegraphics[clip,width=0.49\columnwidth]{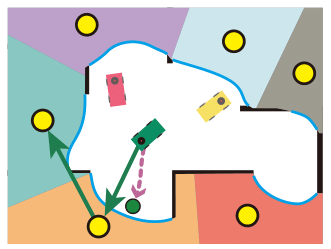}\label{hierarchical1}}
    \subfloat[]{\includegraphics[clip,width=0.49\columnwidth]{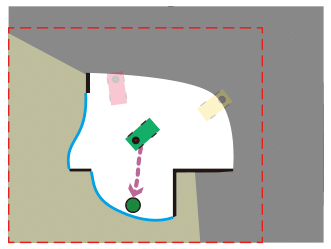}\label{hierarchical2}}

    \subfloat[]{\includegraphics[clip,width=0.33\columnwidth]{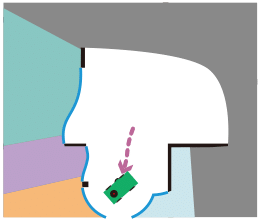}\label{hierarchical3}}
    \subfloat[]{\includegraphics[clip,width=0.33\columnwidth]{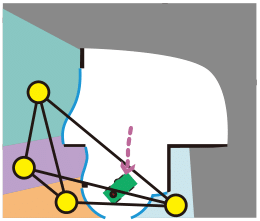}\label{hierarchical4}}
    \subfloat[]{\includegraphics[clip,width=0.33\columnwidth]{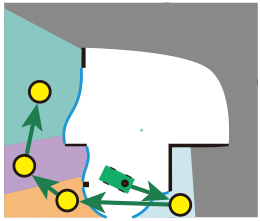}\label{hierarchical5}}

    \subfloat{\includegraphics[clip,width=0.97\columnwidth]{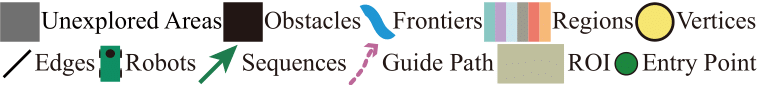}}
    \caption{The Hierarchical Framework. The center node assigns subtasks to the client nodes, as shown in~\ref{hierarchical1}. In~\ref{hierarchical2}, a client node receives the task and transfer along the guide path to the first region.~\ref{hierarchical3},~\ref{hierarchical4}, and~\ref{hierarchical5} are the map within the red dashed box in~\ref{hierarchical2}. The client node partitions the unexplored areas covered by ROI into regions, as shown in~\ref{hierarchical3}. It then constructs a RegionGraph in~\ref{hierarchical4} and solves the TSP to generate the optimal visiting sequence before navigating to the first target, as shown in~\ref{hierarchical5}.}\label{fig:hierarchical}
    \vspace{-1em}
\end{figure}

Each node operates within an ROI that defines the scope of the current exploration task.
The ROI of the center node is typically the entire environment, while the ROI of the client node is given in the received subtask.
The process of hierarchical framework is shown in Fig.~\ref{fig:hierarchical}. The overall exploration task is hierarchically performed.
The center node is responsible for global decision-making by receiving the map and the position of robots and sending the merged global map to all client nodes.
Following the method outlined in~\ref{sec:RegionGraph}, the center node constructs a RegionGraph, obtaining optimal visiting sequences and guide paths for all client nodes.
Subtasks are generated based on these sequences and then dispatched to client nodes.

Upon receiving the subtask, client nodes proceed to explore the areas covered by ROI.
They follow the guide path in the subtasks to reach the first region and then perform an exploration algorithm similar to that of the center node.
The difference is twofold: the VRP for client nodes can be simplified to a TSP since each client node only performs a single robot, and the client node sends the first region of the sequence as the navigation goal to the navigation module instead of a subtask.

\begin{figure}[t]
    \vspace{0.43em}
    \centering
    \includegraphics[width=\columnwidth]{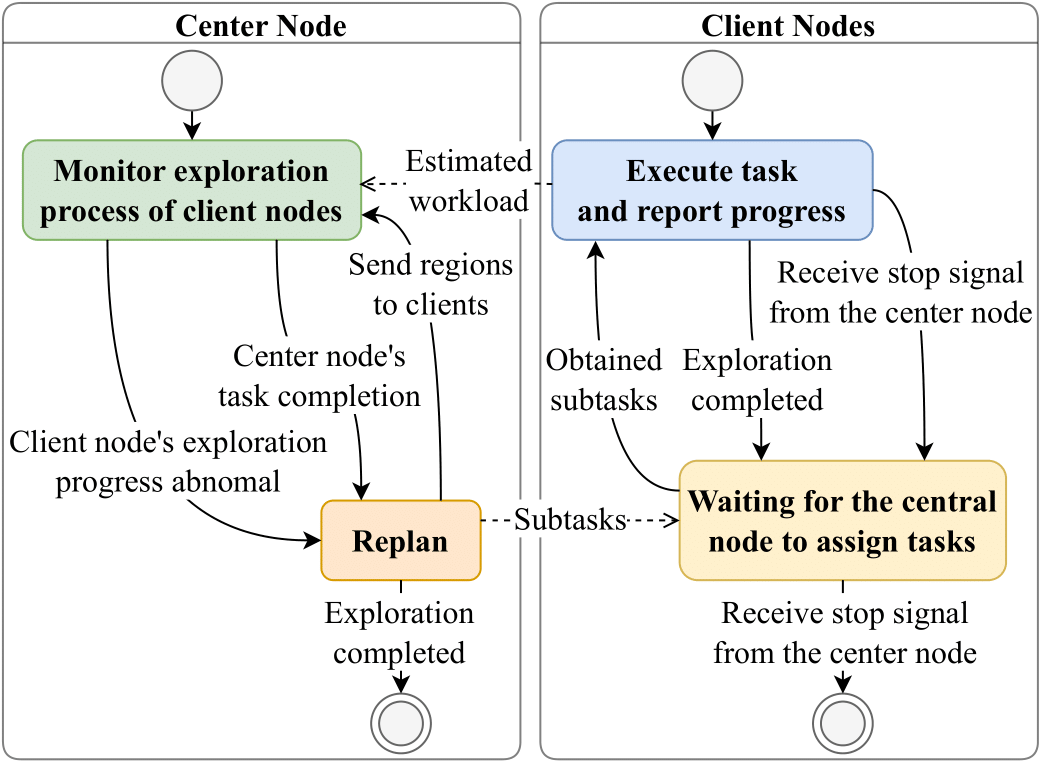}
    \caption{The State Diagram of the Hierarchical Framework}\label{fig:stat_diagram}
    \vspace{-1em}
\end{figure}
\begin{figure}[t]
    \centering
    \subfloat[Empty]{\includegraphics[height=0.2\linewidth,angle=90,origin=c]{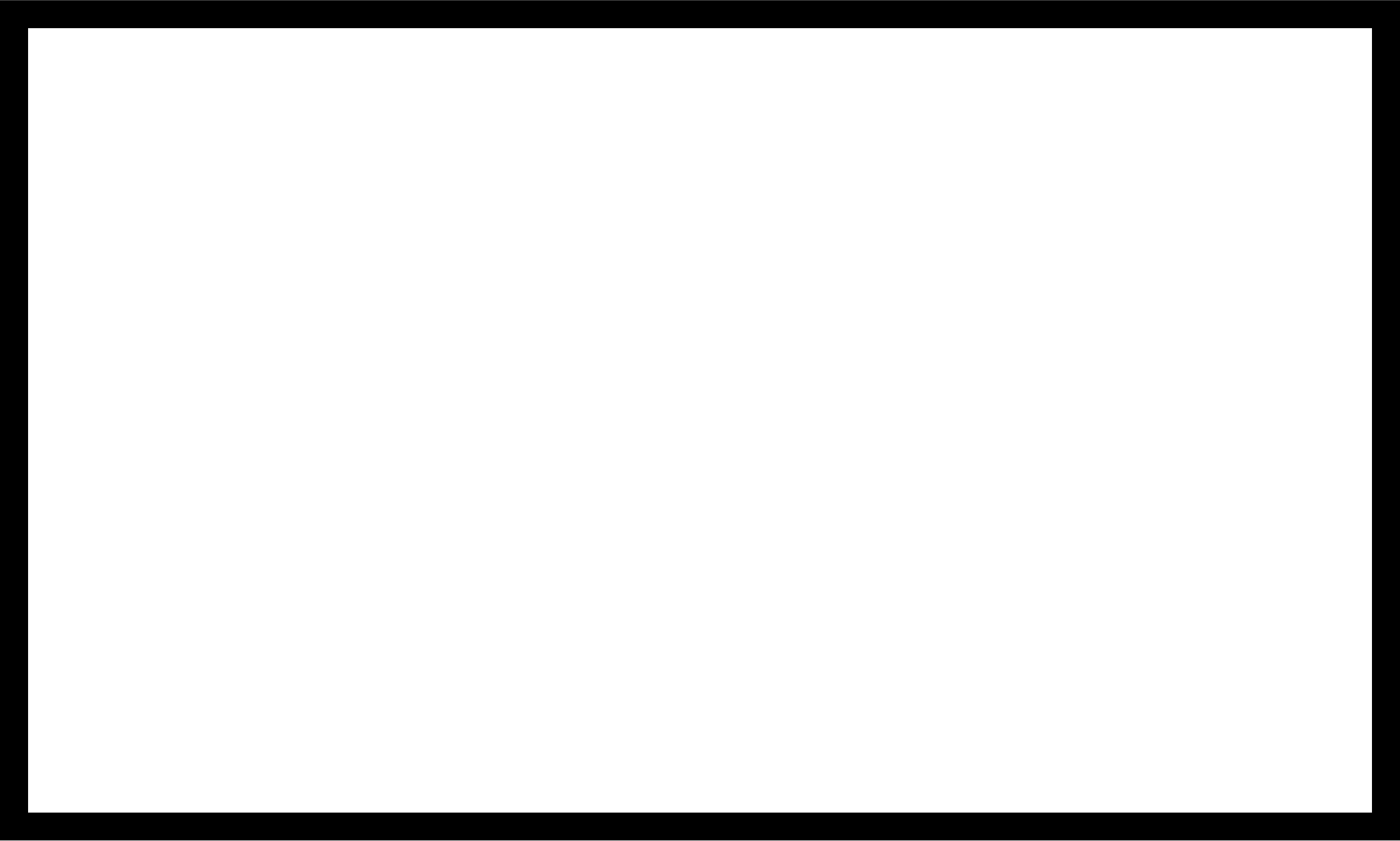}}
    \hspace{0.02\linewidth}
    \subfloat[Grid]{\includegraphics[height=0.2\linewidth,angle=90,origin=c]{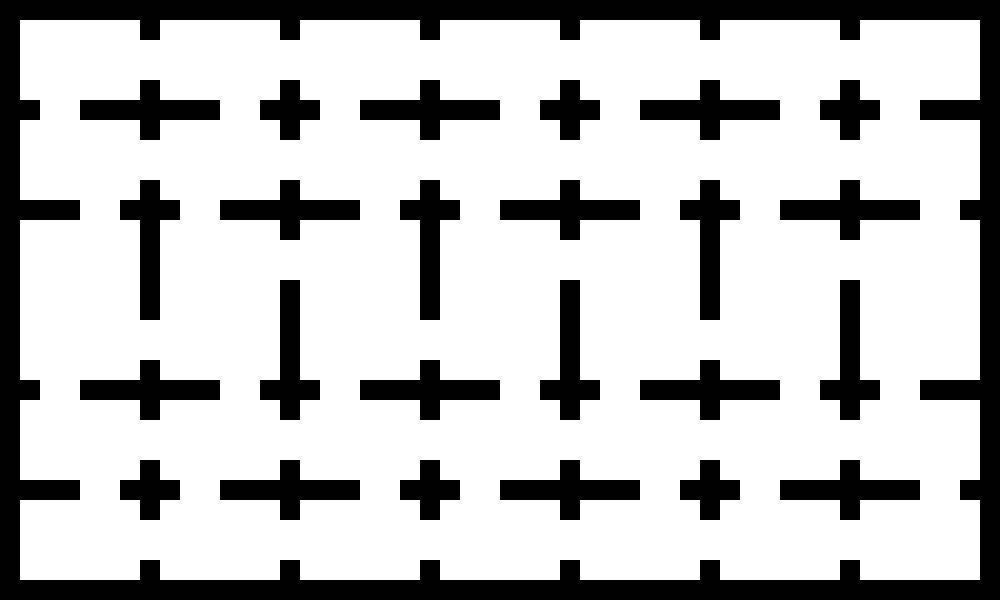}}
    \hspace{0.02\linewidth}
    \subfloat[Random]{\includegraphics[height=0.2\linewidth,angle=90,origin=c]{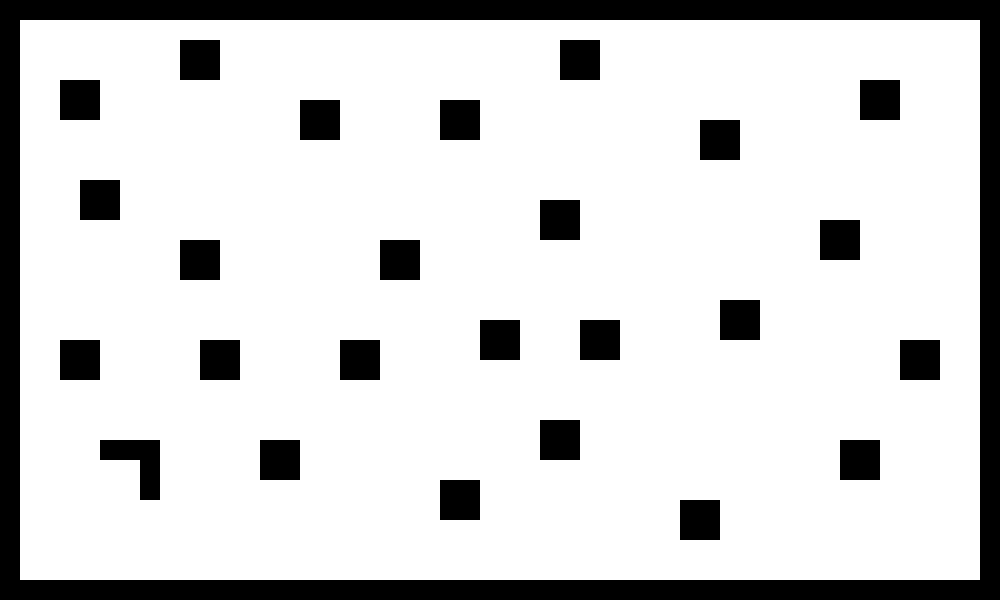}}
    \hspace{0.02\linewidth}
    \subfloat[Campus]{\includegraphics[height=0.2\linewidth,angle=90,origin=c]{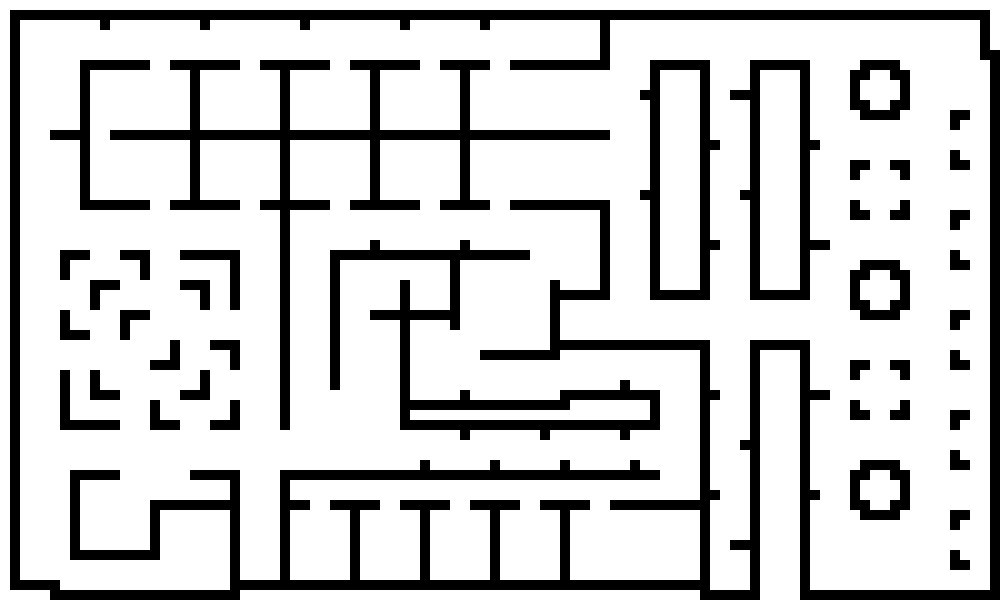}}
    \caption{Environments Used in Simulation ($\SI{50}{\meter}\times \SI{30}{\meter}$).}\label{fig:simulation_env}
    \vspace{-1em}
\end{figure}
\subsubsection{Subtask}
Subtasks refer to tasks dispatched from the center node to client nodes, they consist of two components:
\begin{itemize}
    \item A ROI: The ROI specifies the area that the client node must explore, typically provided as a mask. The client node plans only in its own ROI. When planning is complete, the center node collects the regions in the optimal visiting sequence as an ROI.
    \item A Guide Path: Subtasks also include a guide path to the first region of the sequence. By following the path determined by the center node, the robot explores in accordance with the globally planned results of the center node.
\end{itemize}

\subsubsection{Replanning Mechanism}
The state diagram is shown in Fig.~\ref{fig:stat_diagram}. When a client node is performing an subtask, it calculates its remaining workload using the workload estimation method provided in (2). The node continuously provides feedback on its progress to the center node. The center node will automatically trigger replanning under the following conditions:
\begin{itemize}
    \item Manual instruction for replanning or starting exploration.
    \item Completion or failure of a subtask.
    \item Detection of abnormal progress of a client node. The center node compares the client's estimated workload, actual workload, and remaining workload against a specific threshold to determine if abnormal progress has occurred, as stated in (7).
          \begin{equation}
              w^k_\text{explored}+\hat{w}^k_\text{remaining}-\hat{w}^k_\text{expected}\ge w_\text{threshold}
          \end{equation}
\end{itemize}

The replanning mechanism of the client node is simple: replanning is triggered as soon as the robot reaches its navigation goal. This means that only the first goal of the sequence is executed by the client node, resulting in a rolling update for the client node's exploration.

\section{EXPERIMENTS}\label{sec:simulation_and_experiments}
\subsection{Simulation}

\begin{figure}[t]
    \vspace{0.4em}
    \centering
    \includegraphics[width=0.9\columnwidth]{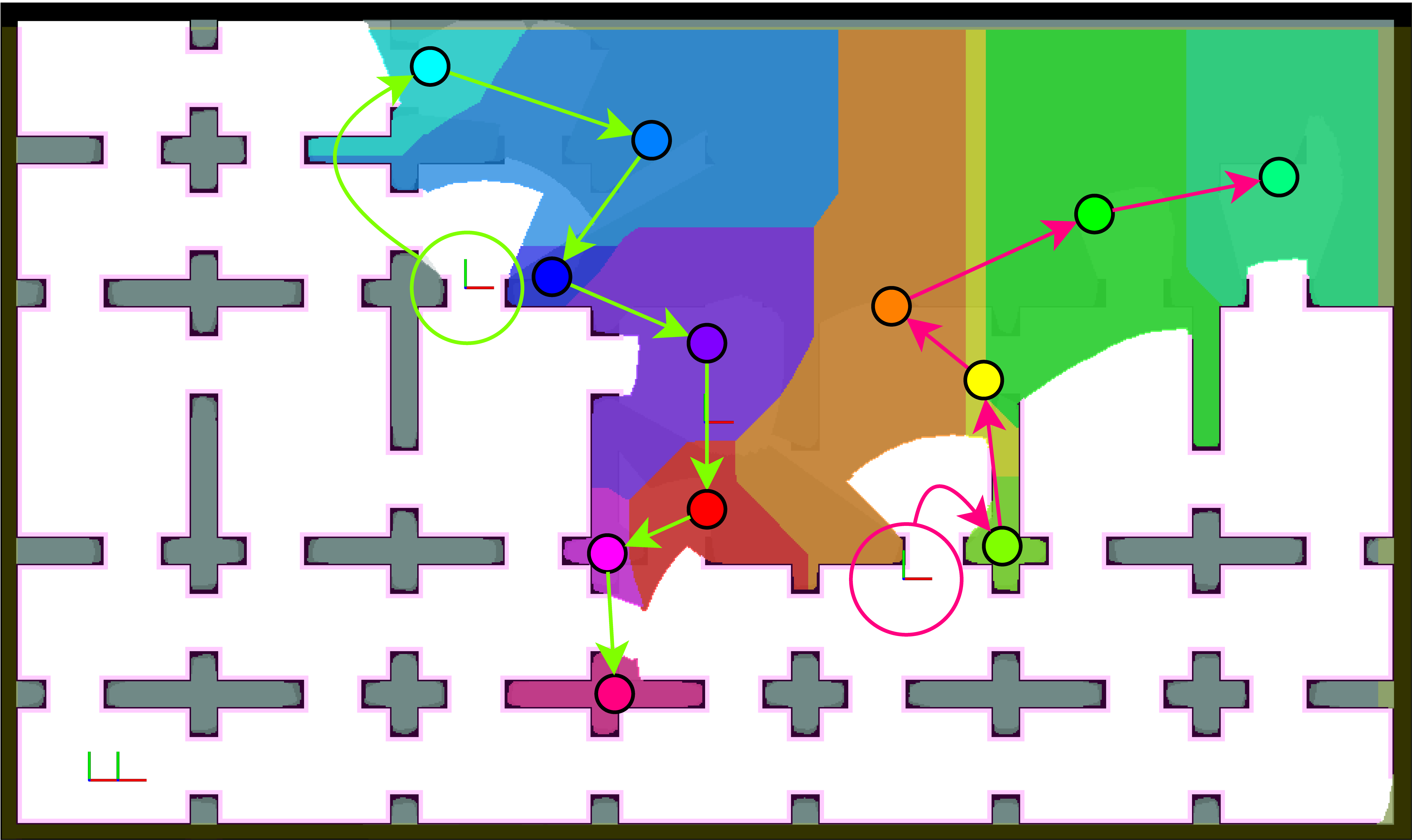}
    \caption{A Task Allocation during Exploration in the Grid Environment Using Two Robots. The figure shows the current position of the robots as the hollow circle connected to arrows. Unexplored regions are represented in different colors. The arrow indicates the optimal visiting sequences.}\label{fig:simulation}
    \vspace{-1em}
\end{figure}
\begin{figure}[t]
    \centering
    \subfloat[Our Method]{%
        \includegraphics[clip,width=0.9\columnwidth]{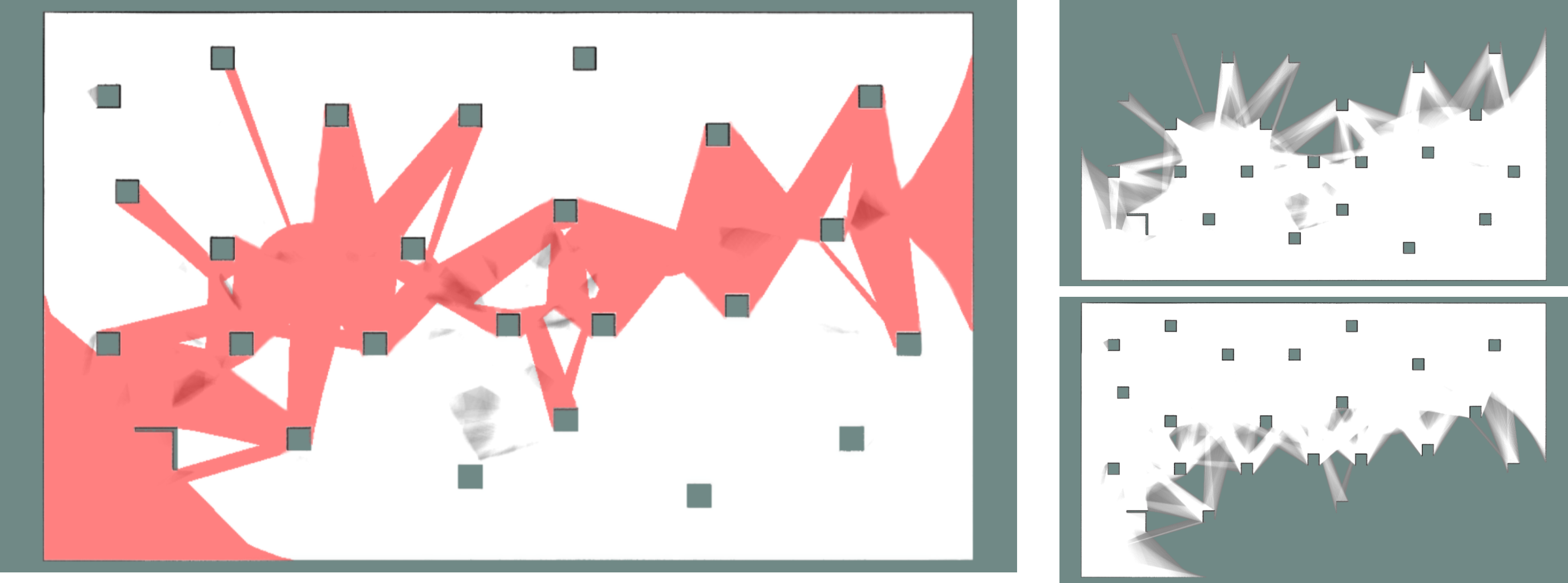}%
    }

    \subfloat[Greedy Method]{%
        \includegraphics[clip,width=0.9\columnwidth]{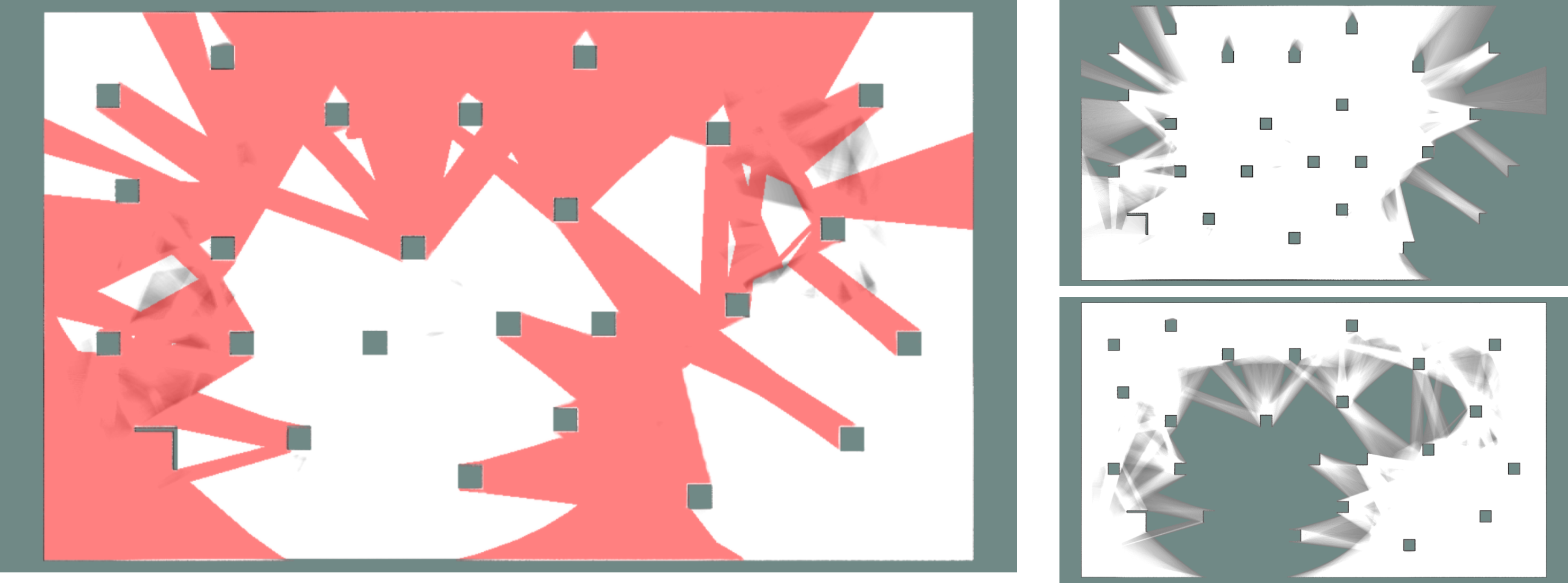}%
    }
    \caption{The Overlap Ratio of the Random Environment Using Two Robots. The left image represents the total explored area, with the red portion indicating overlapping regions. The right image shows areas explored by each of the two robots.}\label{fig:overlap_ratio}
    \vspace{-1em}
\end{figure}
\begin{figure}[t]
    \vspace{0.2em}
    \centering
    \includegraphics[clip,width=0.7\columnwidth]{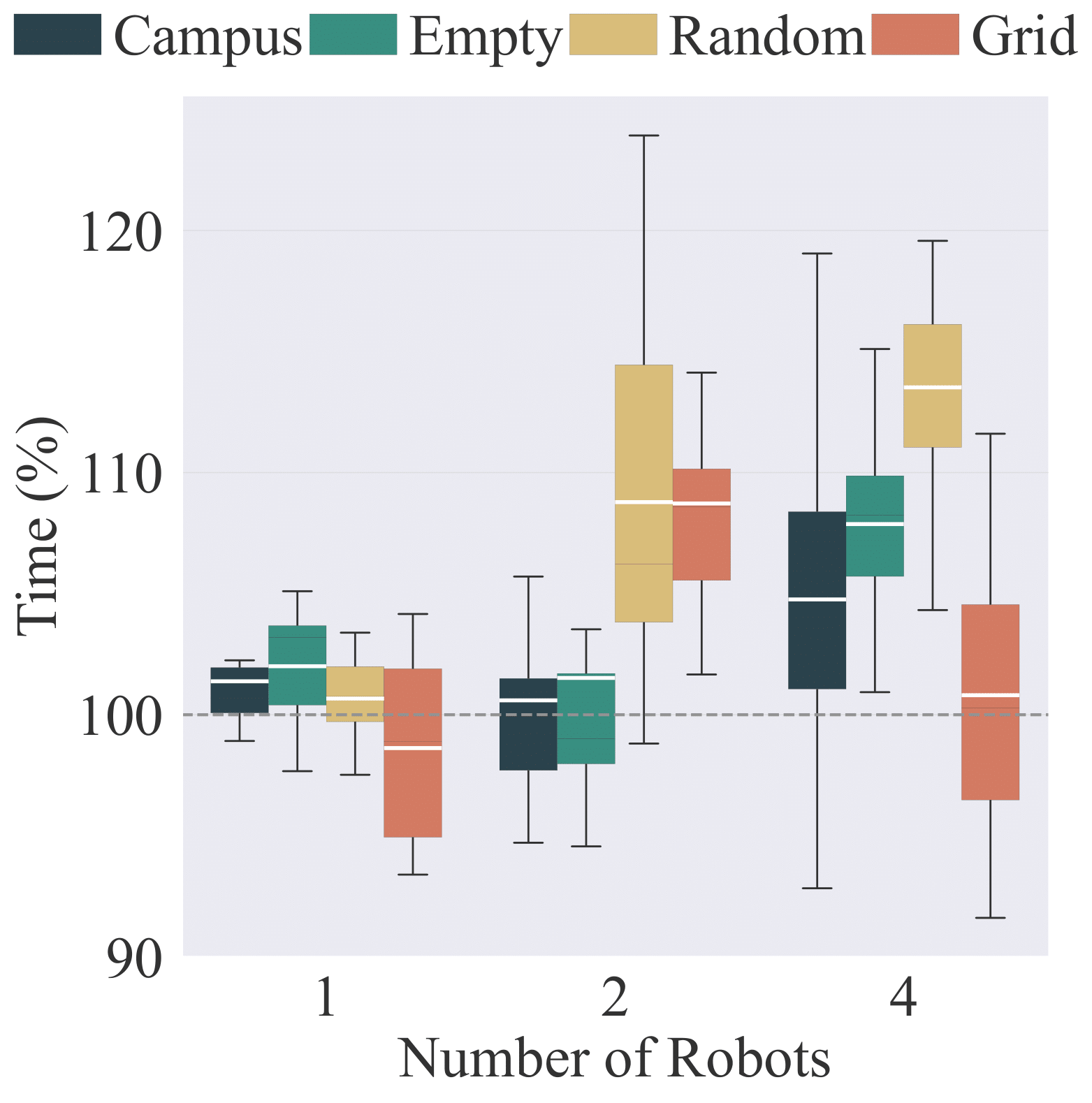}%
    \caption{Exploration Time of Hierarchical Approach versus Centered Approach without VRP Computation Time. The former is set as the baseline (100\%), and a larger latter indicates a longer exploration time. }\label{fig:total_moving_time}
    \vspace{-1em}
\end{figure}
\begin{figure*}[t]
    \vspace{-0.5em}
    \centering
    \subfloat[AET, 1 robot]{
        \includegraphics[clip,width=0.24\linewidth]{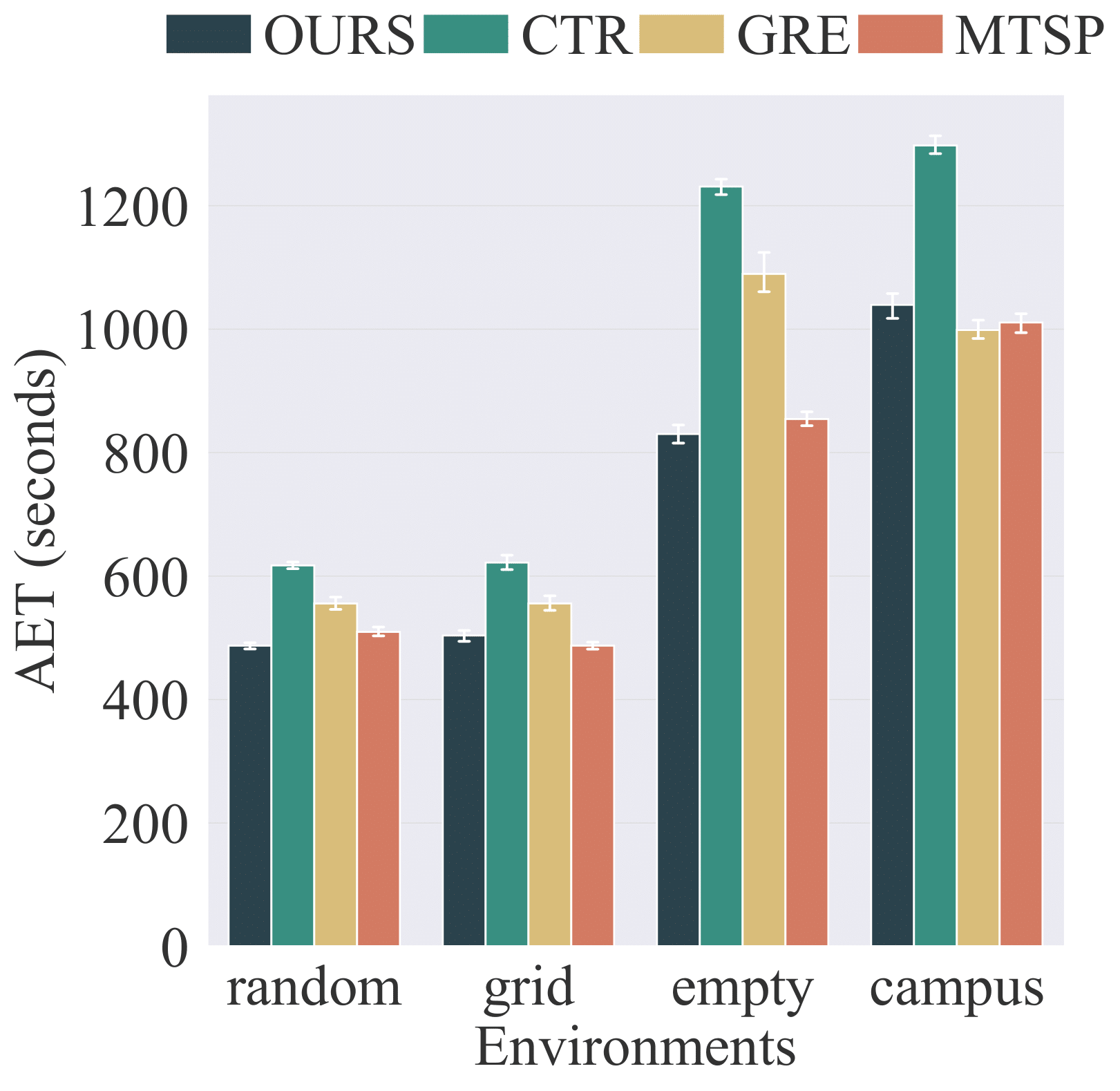}
    }
    \subfloat[AET, 2 robots]{
        \includegraphics[clip,width=0.235\linewidth]{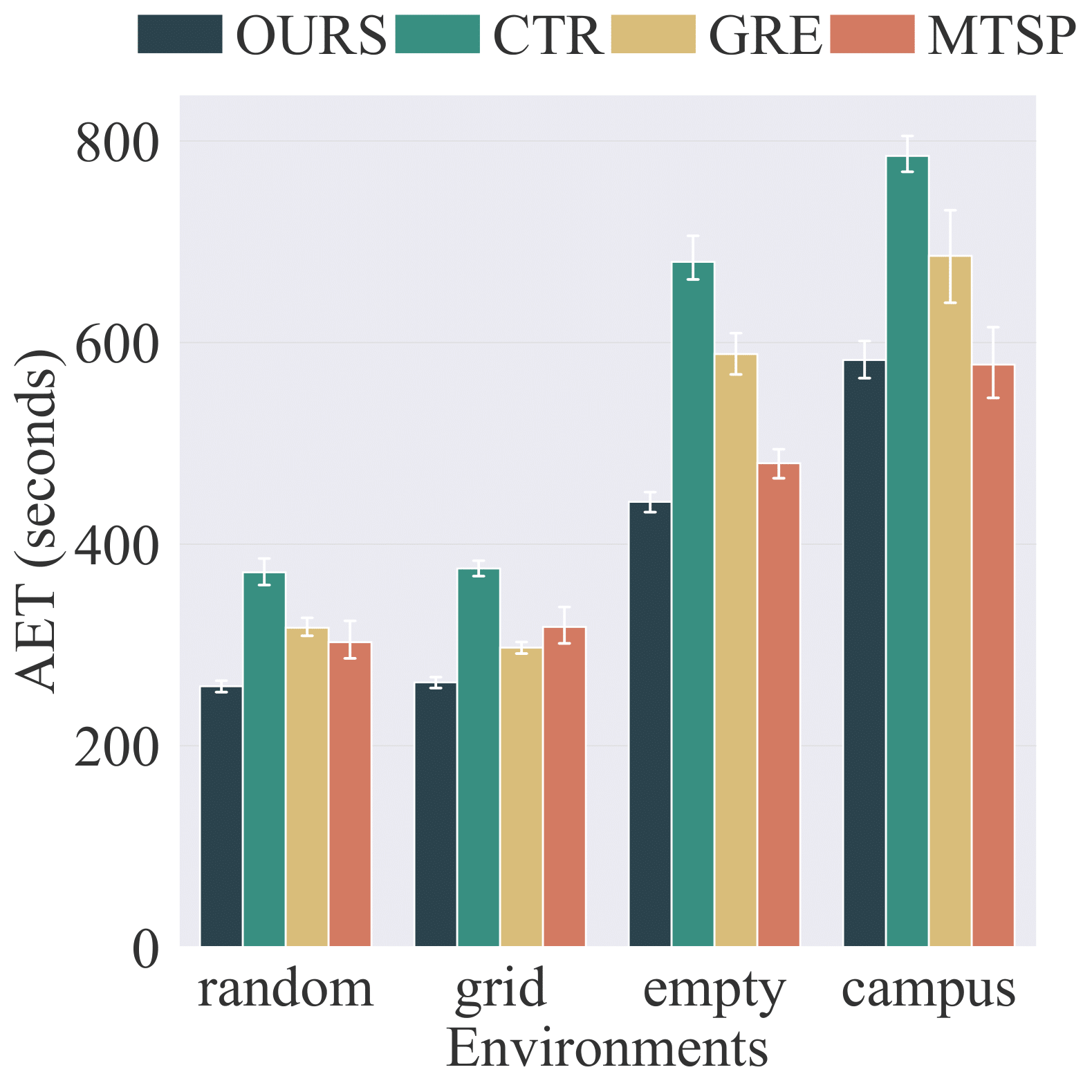}
    }
    \subfloat[AET, 4 robots]{
        \includegraphics[clip,width=0.235\linewidth]{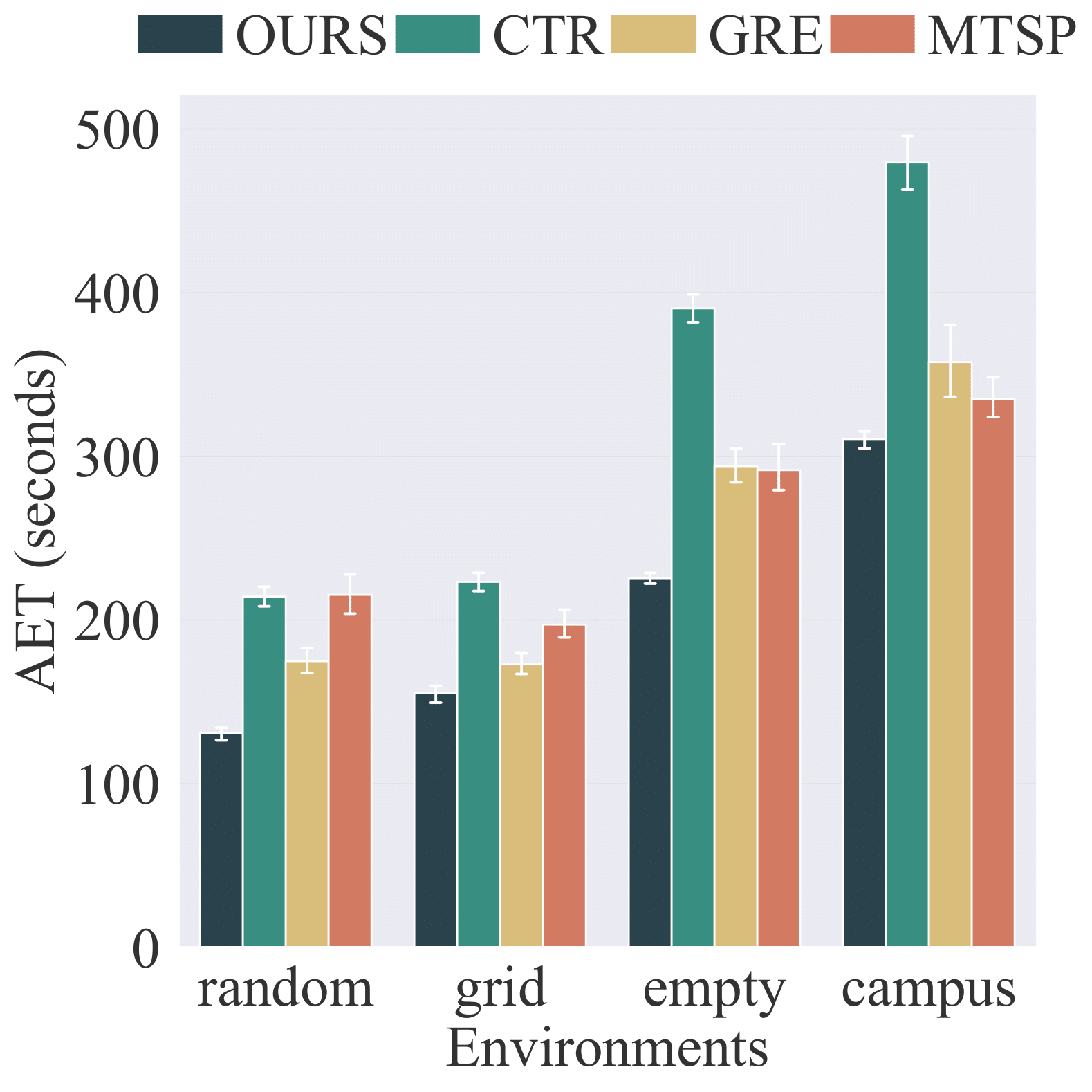}
    }
    \subfloat[ATD, 1 robot]{
        \includegraphics[clip,width=0.235\linewidth]{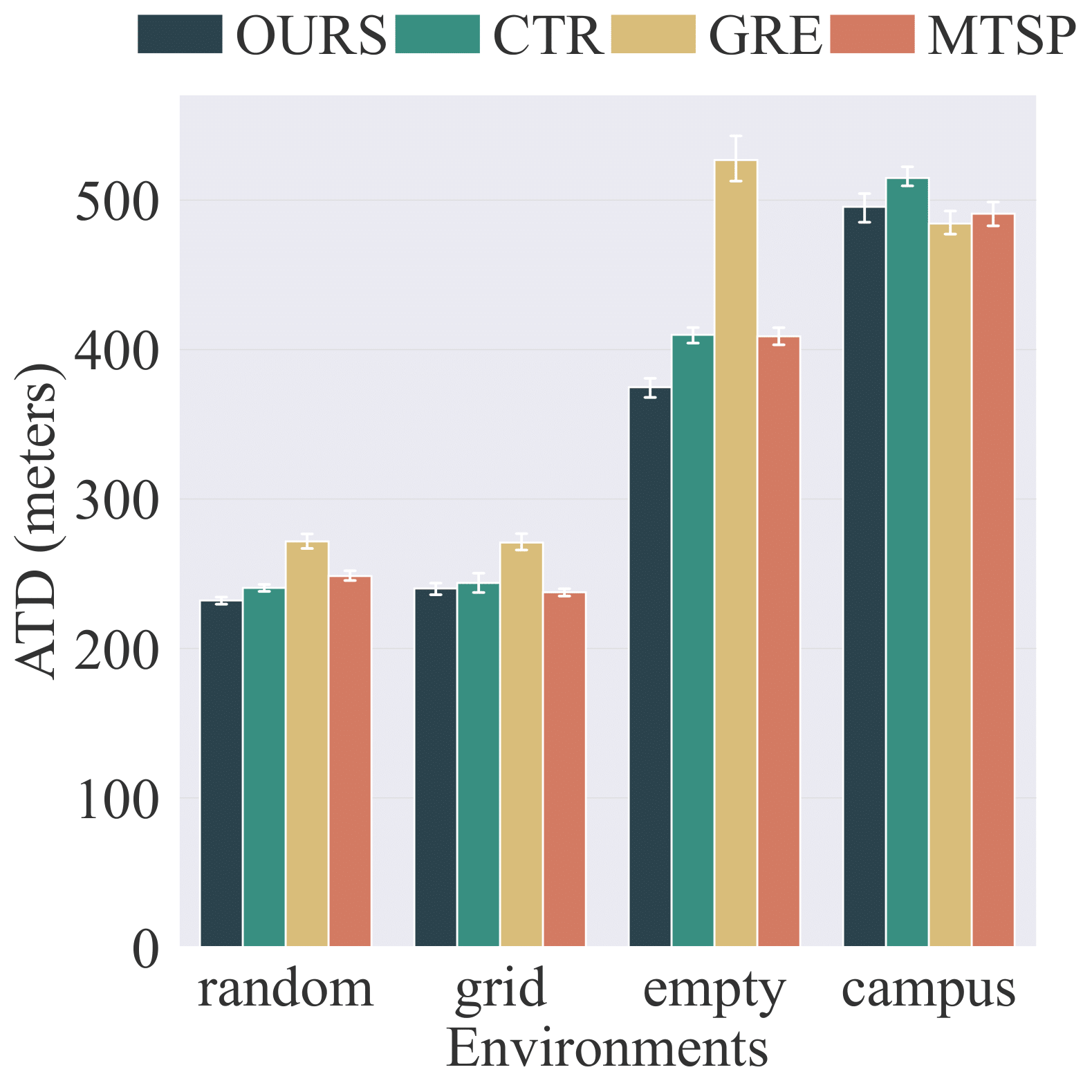}
    }
    % \hspace{0.01em}
    \vspace{-1em}
    \subfloat[ATD, 2 robots]{
        \includegraphics[clip,width=0.235\linewidth]{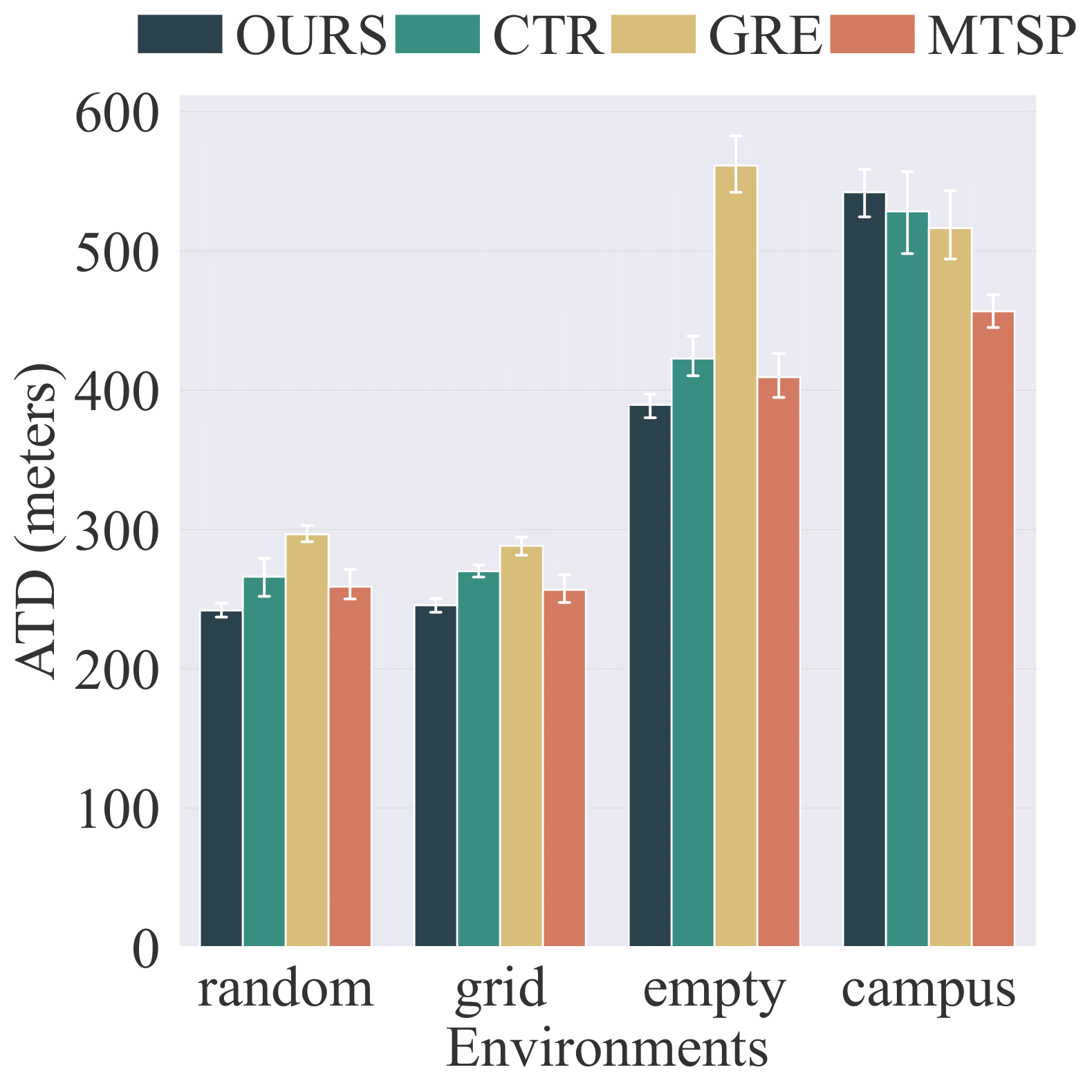}
    }
    \subfloat[ATD, 4 robots]{
        \includegraphics[clip,width=0.235\linewidth]{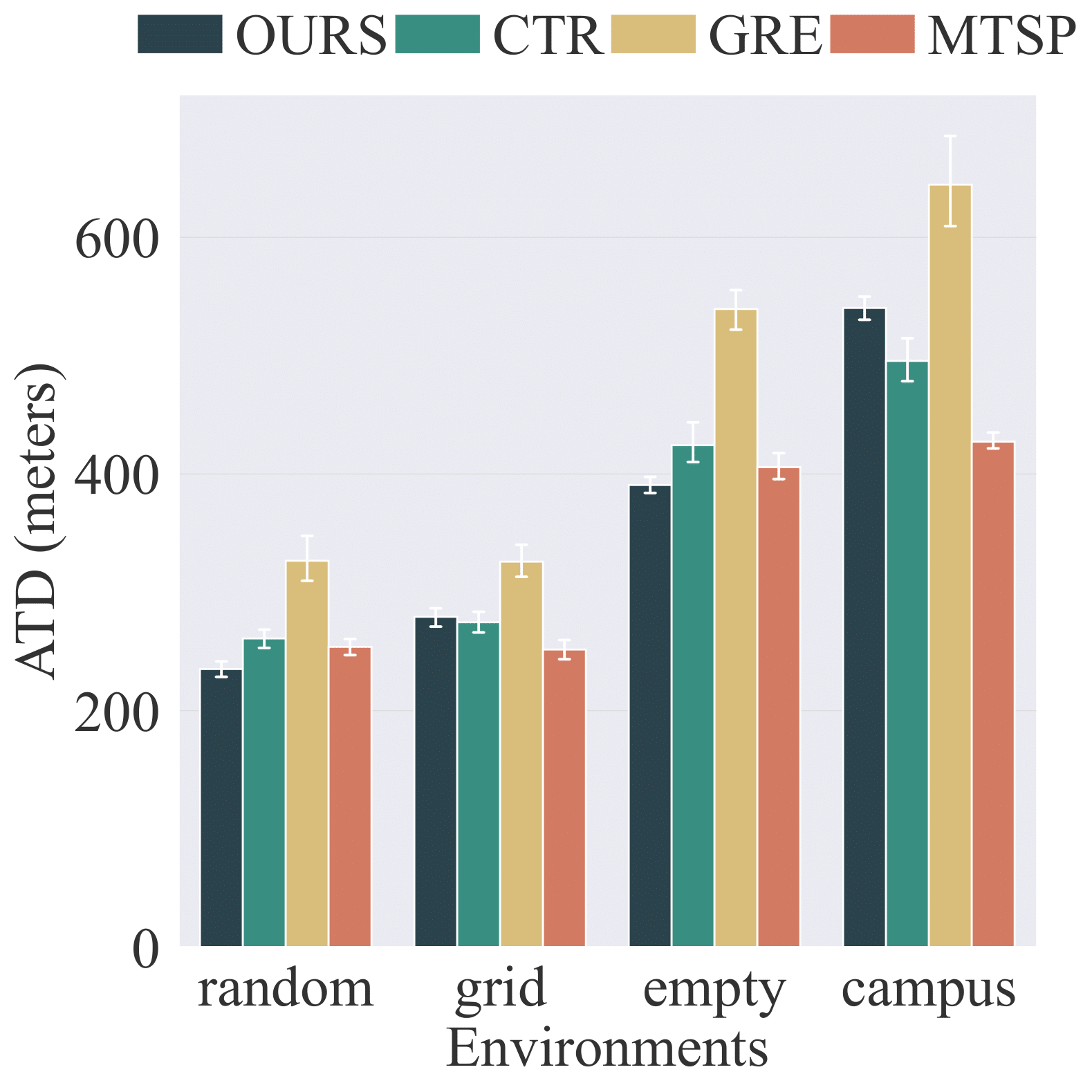}
    }
    \subfloat[AOR, 2 robots]{
        \includegraphics[clip,width=0.235\linewidth]{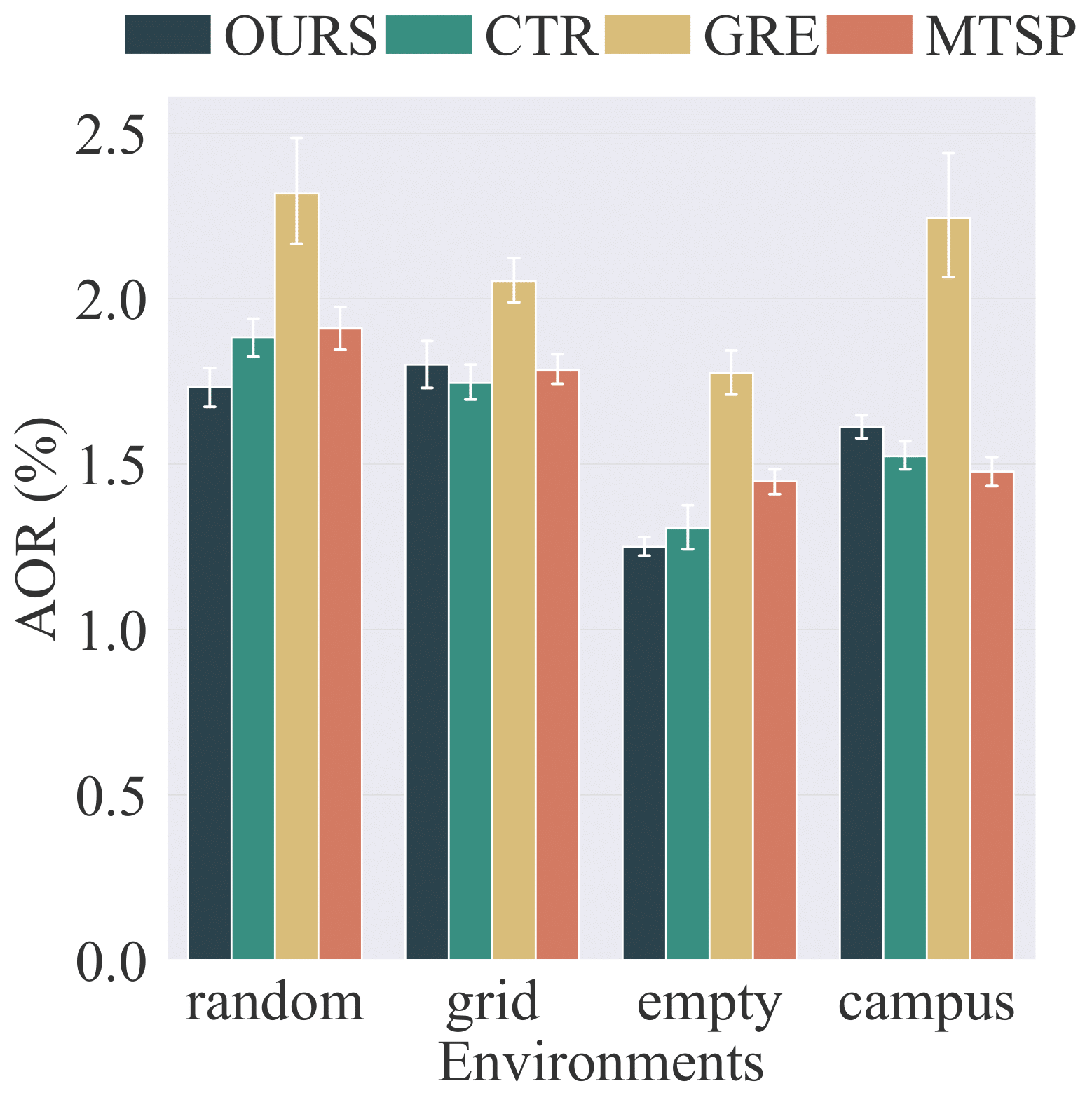}
    }
    \subfloat[AOR, 4 robots]{
        \includegraphics[clip,width=0.235\linewidth]{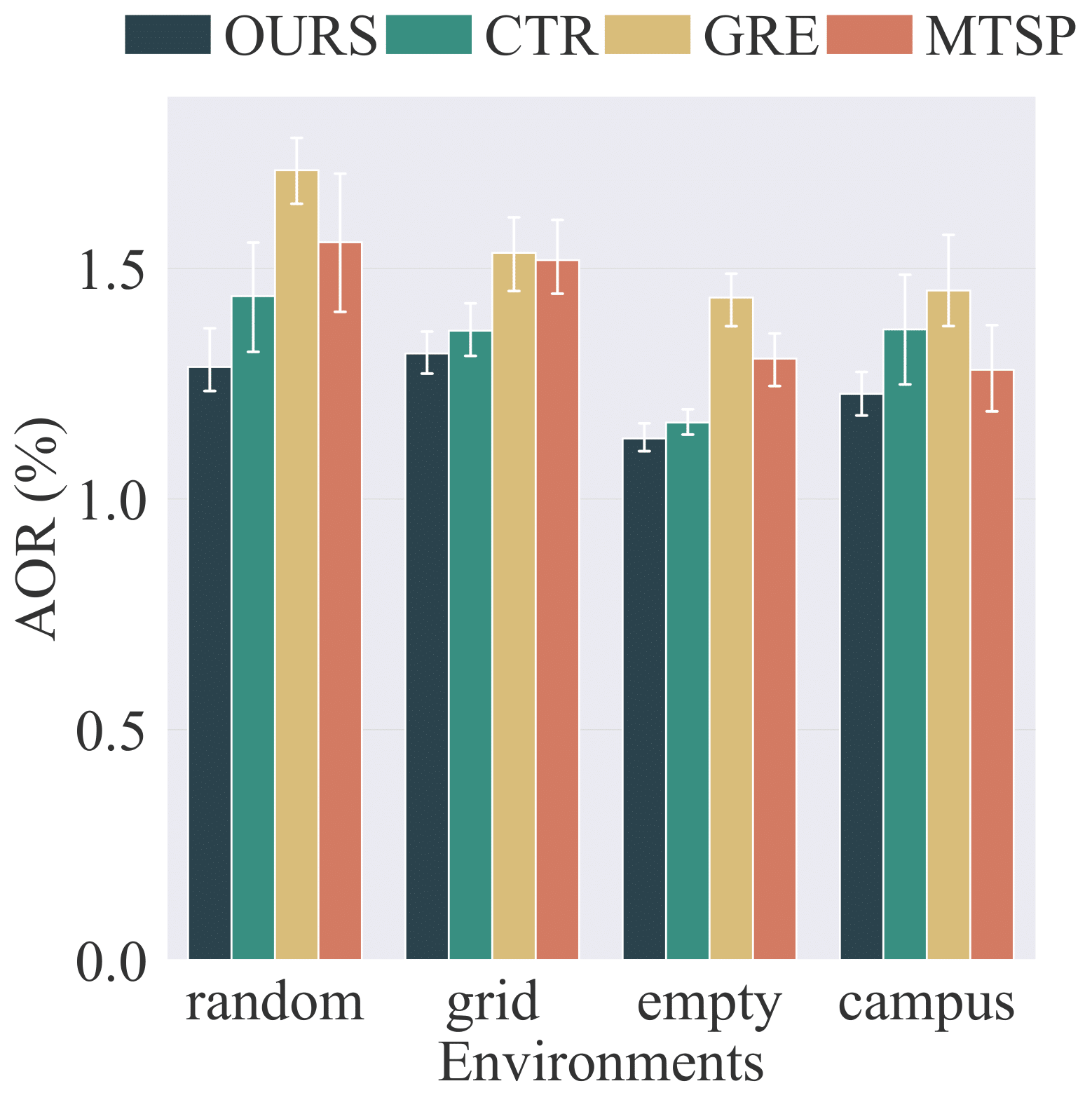}
    }
    \caption{Metrics of Simulation Results}\label{fig:simulation_results}
    \vspace{-1em}
\end{figure*}

\begin{figure*}[t]
    \centering
    \includegraphics[clip,width=0.97\linewidth]{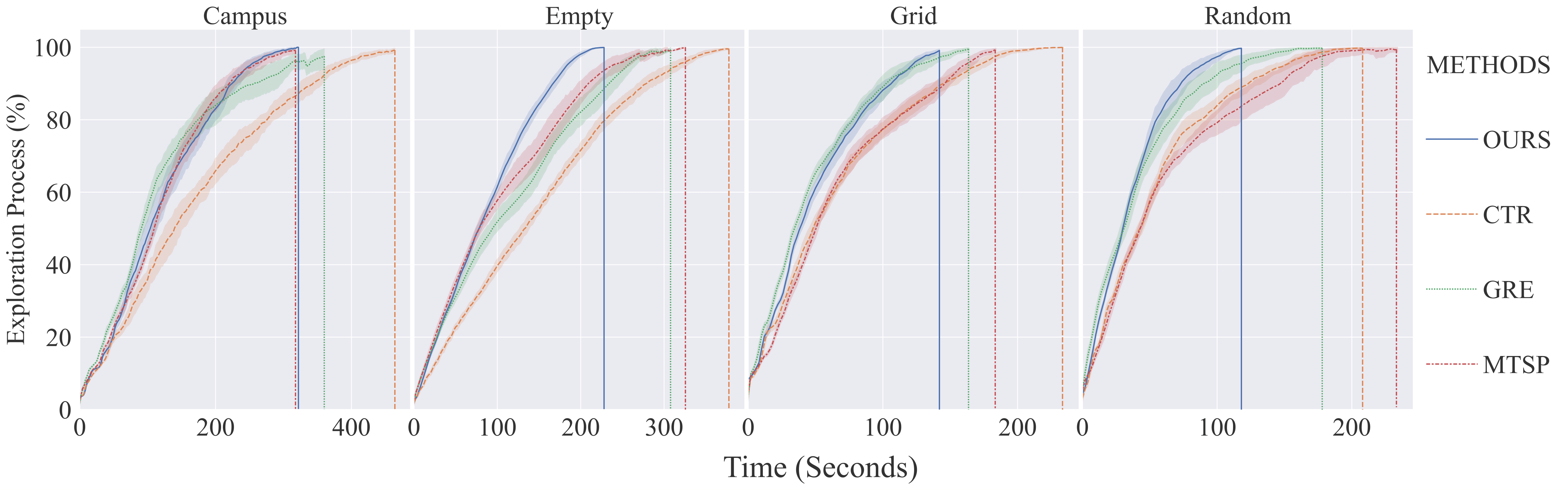}
    \caption{Exploration Process of Simulation in Various Environments Using Four Robots}\label{fig:exploration_process}
    \vspace{-1em}
\end{figure*}

\subsubsection{Configuration}
The proposed method is based on the Robot Operation System 2 (ROS2). The algorithm's effectiveness is verified in a 2D simulator. The experiments are conducted in a ROS2 Galactic-based Docker container. The robot is configured with a speed of  $\SI{1}{\meter/\second}$ and an effective radar sensor radius of $\SI{6}{\meter}$. The comparison experiments are performed using the following method:
\begin{itemize}
    \item Centralized method (CTR): This method is the same as our method except the hierarchical framework. It serves as a comparison to evaluate the effectiveness of the hierarchical framework.
    \item Multi-TSP (MTSP): A global decision-making method using Voronoi segmentation based on the position of robots and assigns partitioned regions to robots, which then use a TSP solver to plan within them.
    \item Greedy (GRE): A greedy method using a utility function for target allocation. The frontier points are ranked in descending order based on the utility function, which takes into account the distance to the target points and the information gain of frontier points. To promote dispersion, the utilities of the nearby assigned points are decreased. The process is repeated until all robots or all frontier points are assigned.
\end{itemize}
The comparison algorithms are centralized structures and replanning is triggered when any robot reaches its goal.  It's worth noting that due to the existence of unobservable areas $\Omega_u$, the final map may still contain some unknown regions, which is an expected outcome of the exploration process.

The analysis of results employs the following three metrics:
\begin{itemize}
    \item Average Exploration Time (AET), which is defined as the time from the start to the completion of exploration. This metric directly reflects the exploration efficiency.
    \item Average Distance Traveled (ADT), which is defined as the average distance all robots traveled during the exploration, This metric can indicates the average energy consumption.
    \item Average Overlap Ratio (AOR), which is defined as the ratio of the numerical sum of the areas explored to the area of the union of explored. This metric reflects how effectively the exploration framework is coordinated.
\end{itemize}
To conduct comprehensive comparisons, simulation is performed on four maps with varying topologies and complexities, as depicted in Fig.~\ref{fig:simulation_env}. Each map involves 1, 2, and 4 robots, and multiple trials are conducted to minimize anomalous data resulting from localization, mapping, and navigation issues. A resut of task allocation is displayed in Fig.~\ref{fig:simulation}.

\subsubsection{Results and Analysis}

The simulation results are presented in Fig.~\ref{fig:simulation_results}. The statistical results demonstrate that our method outperforms 20\% more than others in terms of AET, ADT, and AOR in most cases involving multiple robots scenarios.

The curve of the explored area over time is shown in Fig.~\ref{fig:exploration_process}. It is evident that during the initial stage of exploration, our method does not demonstrate a significant advantage compared to others. However, as the exploration progresses, other methods may leave small, distant areas unexplored, resulting in decrease in exploration efficiency over time. In contrast, ours maintains high exploration efficiency and ultimately completes the exploration task faster.

The overlap ratio is shown in Fig.~\ref{fig:overlap_ratio}, where the red part indicates areas explored by more than one robots. It is evident that our method results in fewer overlapping areas, reflecting its advantage in multi-robot coordination.

However, it is important to note that the total distance traveled and the exploration time may not always be strictly positively correlated in certain scenarios, which is because other methods do not consider the area information of unexplored regions during task allocation, only decides based on the distance of target points. This leads to robots that are farther from target points being left without tasks.

The comparison between our method and the CTR without VRP computation time is shown in Fig.~\ref{fig:total_moving_time}.
Our method is more efficient than the CTR, and it does not compromise the global planning effect, even though it reduces the frequency of global decision-making.
Additionally, our exploration framework achieves non-stop replanning, eliminating the computational time required for centralized method. This improves exploration efficiency by avoiding robots remaining stationary due to lengthy calculations during decision-making.

\subsection{Real-World Experiments}
\begin{figure}[t]
    \centering
    \subfloat[]{
        \includegraphics[clip,width=0.50\columnwidth]{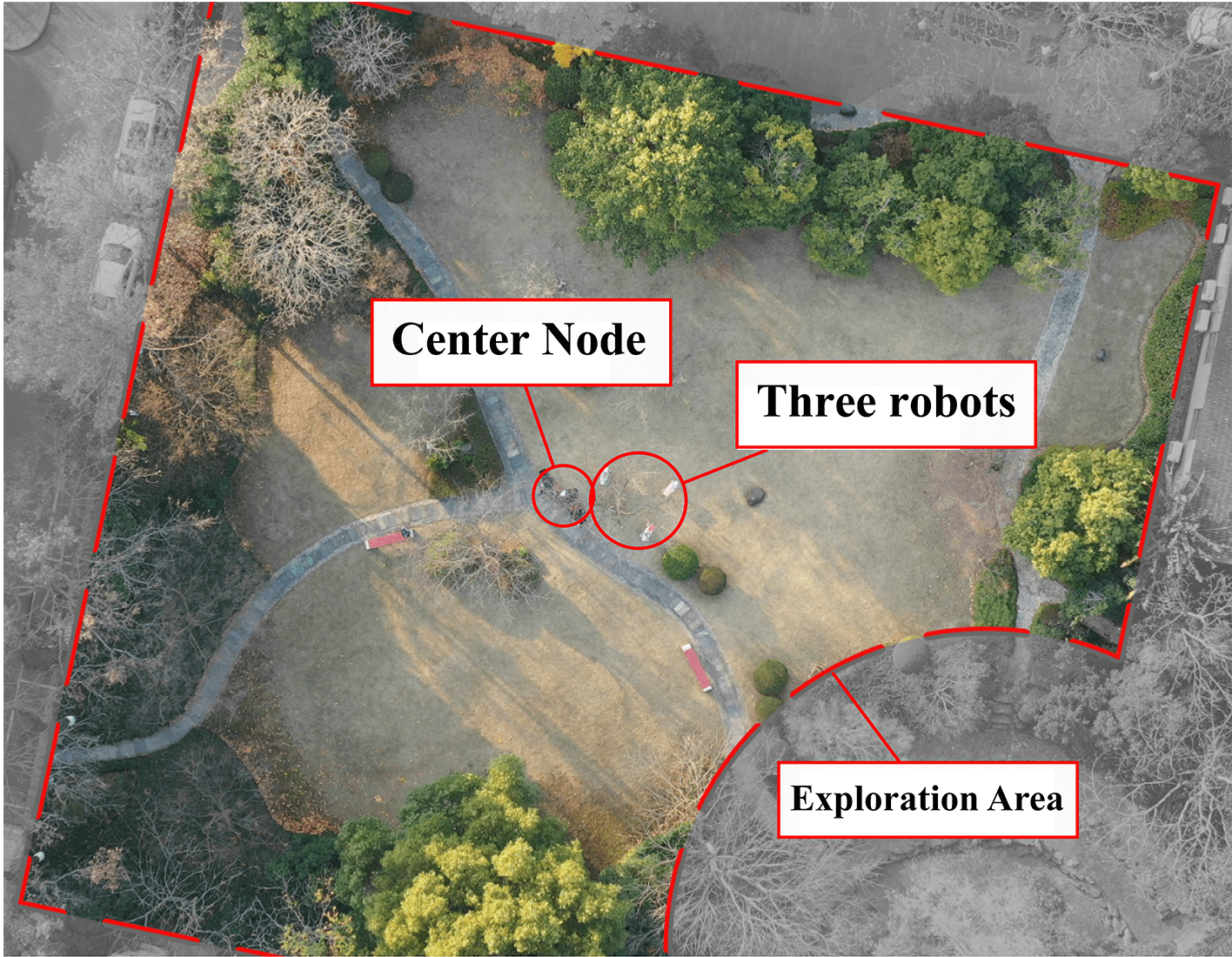}\label{fig:experiment_environment}}
    \hspace{0.01\linewidth}
    \subfloat[]{\includegraphics[clip,width=0.382\columnwidth]{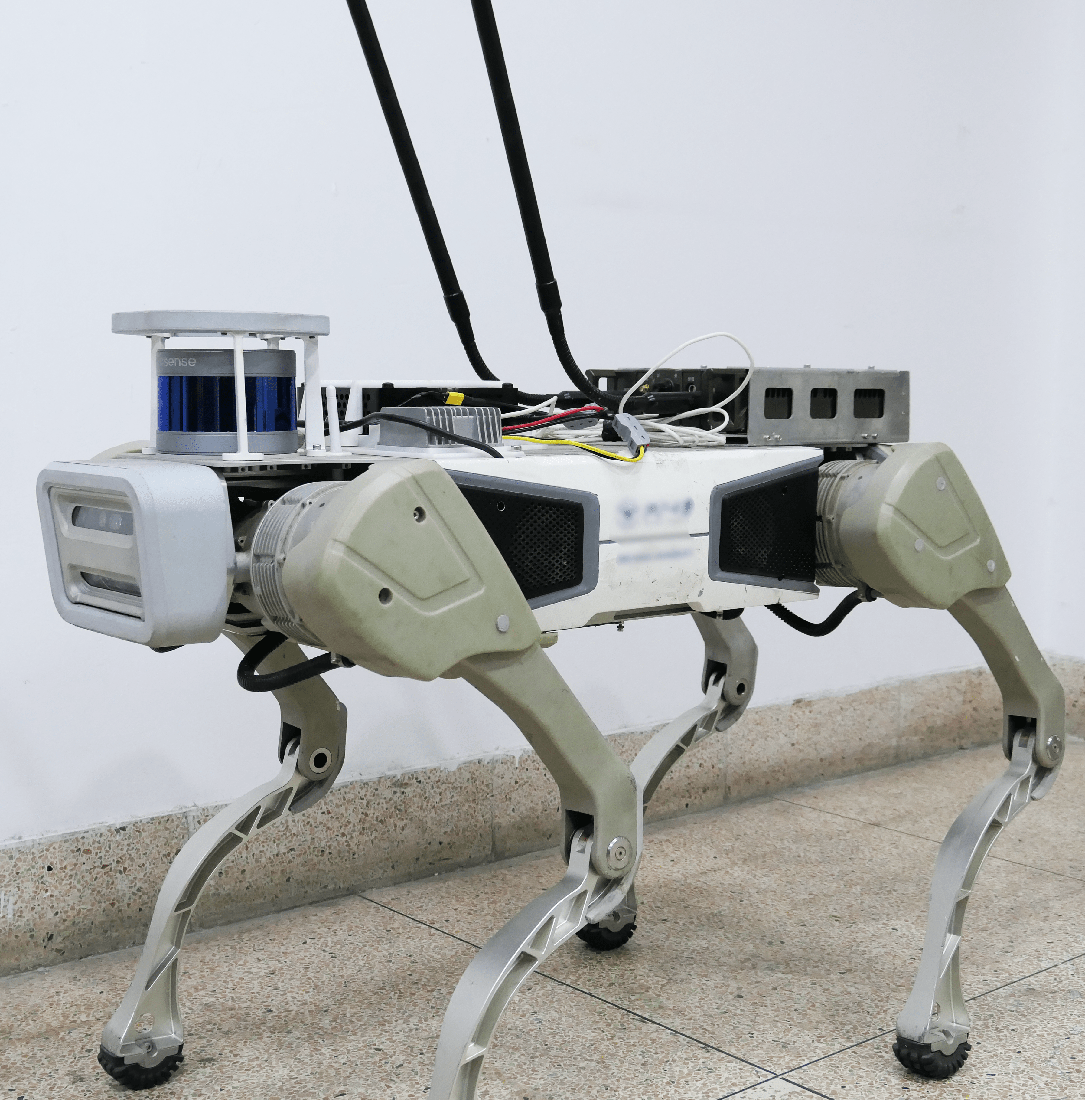}\label{fig:jueying}
    }
    \caption{(a) Environment of Experiment (b) Jueying Robot}
    \vspace{-1.5em}
\end{figure}
The experiment are conducted on three quadrupedal robots called Jueying X20, as shown in Fig.~\ref{fig:jueying}.Our method is decoupled from the implemented platform. Legged robots are chosen as the experimental platform due to their ability to move stably in complex outdoor scenarios. The robot has a speed of $\SI{1}{\meter/\second}$ and is powered by a NUC with an AMD R9 CPU and 16GiB DDR4 RAM. Robots are equipped with a 16-line Robosense LiDAR, with an effective radius of $\SI{6}{\meter}$ for consistency with the simulation. The framework is deployed on the NUC of these robots.

In practical applications, it is often necessary to determine the relative poses of robots during exploration. This requires simultaneous multi-robot map fusion and relative pose estimation, which is not investigated in detail here. Instead, a global localization solution is utilized to establish the relative poses of robots and their coordinate systems. All robots use a point cloud map and an HDL-localization module for localization. The Navigation2 module is used for navigation during the experiments. The experiment is conducted in a garden with an area of \SI{1054}{\meter\squared}, as depicted in Fig.~\ref{fig:experiment_environment}. The designated exploration zone is indicated by the red box. Three robots are used to explore the area, with the center node deployed on a PC located at the center of the garden.
Figure~\ref{fig:frontpage} shows the results of the allocation and the exploration map after the experiment is completed. The exploration is completed within 242 seconds using three robots.

\section{CONCLUSION}\label{sec:conclusion}

We proposed a hierarchical framework leveraging RegionGraph to improve exploration efficiency by preserving global spatial information and enabling asynchronous operations.  Further research is needed to extend the method to three-dimensional environments.

\bibliographystyle{IEEEtran} % use IEEEtran.bst style
\bibliography{./IEEEabrv,./ref}

\begin{thebibliography}{10}
\providecommand{\url}[1]{#1}
\csname url@rmstyle\endcsname
\providecommand{\newblock}{\relax}
\providecommand{\bibinfo}[2]{#2}
\providecommand\BIBentrySTDinterwordspacing{\spaceskip=0pt\relax}
\providecommand\BIBentryALTinterwordstretchfactor{4}
\providecommand\BIBentryALTinterwordspacing{\spaceskip=\fontdimen2\font plus
\BIBentryALTinterwordstretchfactor\fontdimen3\font minus \fontdimen4\font\relax}
\providecommand\BIBforeignlanguage[2]{{%
\expandafter\ifx\csname l@#1\endcsname\relax
\typeout{** WARNING: IEEEtran.bst: No hyphenation pattern has been}%
\typeout{** loaded for the language `#1'. Using the pattern for}%
\typeout{** the default language instead.}%
\else
\language=\csname l@#1\endcsname
\fi
#2}}

\bibitem{yamauchi_frontier-based_1997}
B.~Yamauchi, ``\BIBforeignlanguage{en}{A frontier-based approach for autonomous exploration},'' in \emph{\BIBforeignlanguage{en}{1997 {IEEE} {International} {Symposium} on {Computational} {Intelligence} in {Robotics} and {Automation} ({CIRA})}}, July 1997, pp. 146--151.

\bibitem{gao2018improved}
W.~Gao, M.~Booker, A.~Adiwahono, M.~Yuan, J.~Wang, and Y.~W. Yun, ``An improved frontier-based approach for autonomous exploration,'' in \emph{2018 15th International Conference on Control, Automation, Robotics and Vision (ICARCV)}.\hskip 1em plus 0.5em minus 0.4em\relax IEEE, 2018, pp. 292--297.

\bibitem{khamis2015multi}
A.~Khamis, A.~Hussein, and A.~Elmogy, ``Multi-robot task allocation: A review of the state-of-the-art,'' \emph{Cooperative robots and sensor networks 2015}, pp. 31--51, 2015.

\bibitem{abonyi2023autonomous}
B.~Abonyi-T{\'o}th and {\'A}.~Nagy, ``Autonomous exploration using a tree structure for goal selection,'' in \emph{2023 IEEE 17th International Symposium on Applied Computational Intelligence and Informatics (SACI)}.\hskip 1em plus 0.5em minus 0.4em\relax IEEE, 2023, pp. 000\,021--000\,026.

\bibitem{bircher_receding_2016}
A.~Bircher, M.~Kamel, K.~Alexis, H.~Oleynikova, and R.~Siegwart, ``\BIBforeignlanguage{zh-CN}{Receding {Horizon} "{Next}-{Best}-{View}" {Planner} for {3D} {Exploration}},'' in \emph{\BIBforeignlanguage{zh-CN}{2016 {IEEE} {International} {Conference} on {Robotics} and {Automation} ({ICRA})}}, May 2016, pp. 1462--1468.

\bibitem{umari_autonomous_2017}
H.~Umari and S.~Mukhopadhyay, ``\BIBforeignlanguage{en}{Autonomous robotic exploration based on multiple rapidly-exploring randomized trees},'' in \emph{\BIBforeignlanguage{en}{2017 {IEEE}/{RSJ} {International} {Conference} on {Intelligent} {Robots} and {Systems} ({IROS})}}, Sept. 2017, pp. 1396--1402.

\bibitem{dang_graph-based_2020}
T.~Dang, M.~Tranzatto, S.~Khattak, F.~Mascarich, K.~Alexis, and M.~Hutter, ``\BIBforeignlanguage{en}{Graph-based {Subterranean} {Exploration} {Path} {Planning} using {Aerial} and {Legged} {Robots}},'' \emph{\BIBforeignlanguage{en}{Journal of Field Robotics}}, Oct. 2020.

\bibitem{dai2020fast}
A.~Dai, S.~Papatheodorou, N.~Funk, D.~Tzoumanikas, and S.~Leutenegger, ``Fast frontier-based information-driven autonomous exploration with an mav,'' in \emph{2020 IEEE international conference on robotics and automation (ICRA)}.\hskip 1em plus 0.5em minus 0.4em\relax IEEE, 2020, pp. 9570--9576.

\bibitem{zhou_fuel_2021}
B.~Zhou, Y.~Zhang, X.~Chen, and S.~Shen, ``\BIBforeignlanguage{English}{{FUEL}: {Fast} {UAV} {Exploration} {Using} {Incremental} {Frontier} {Structure} and {Hierarchical} {Planning}},'' \emph{\BIBforeignlanguage{English}{IEEE Robotics and Automation Letters}}, vol.~6, no.~2, pp. 779--786, Apr. 2021.

\bibitem{zhang2021multi}
Y.~Zhang, Y.~Zhang, and X.~Li, ``Multi-robot exploration system in unknown environment based on submap,'' in \emph{2021 4th International Conference on Robotics, Control and Automation Engineering (RCAE)}.\hskip 1em plus 0.5em minus 0.4em\relax IEEE, 2021, pp. 256--260.

\bibitem{solanas2004coordinated}
A.~Solanas and M.~A. Garcia, ``Coordinated multi-robot exploration through unsupervised clustering of unknown space,'' in \emph{2004 IEEE/RSJ International Conference on Intelligent Robots and Systems (IROS)(IEEE Cat. No. 04CH37566)}, vol.~1.\hskip 1em plus 0.5em minus 0.4em\relax IEEE, 2004, pp. 717--721.

\bibitem{bi2023cure}
Q.~Bi, X.~Zhang, J.~Wen, Z.~Pan, S.~Zhang, R.~Wang, and J.~Yuan, ``Cure: A hierarchical framework for multi-robot autonomous exploration inspired by centroids of unknown regions,'' \emph{IEEE Transactions on Automation Science and Engineering}, no.~99, pp. 1--14, 2023.

\bibitem{mannucci_autonomous_2018}
A.~Mannucci, S.~Nardi, and L.~Pallottino, ``\BIBforeignlanguage{en}{Autonomous {3D} {Exploration} of {Large} {Areas}: {A} {Cooperative} {Frontier}-{Based} {Approach}},'' in \emph{\BIBforeignlanguage{en}{Modelling and {Simulation} for {Autonomous} {Systems}}}, ser. Lecture {Notes} in {Computer} {Science}, J.~Mazal, Ed.\hskip 1em plus 0.5em minus 0.4em\relax Cham: Springer International Publishing, 2018, pp. 18--39.

\bibitem{burgard_coordinated_2005}
W.~Burgard, M.~Moors, C.~Stachniss, and F.~Schneider, ``\BIBforeignlanguage{en}{Coordinated multi-robot exploration},'' \emph{\BIBforeignlanguage{en}{IEEE Transactions on Robotics}}, vol.~21, no.~3, pp. 376--386, June 2005.

\bibitem{yamauchi1998frontier}
B.~Yamauchi, ``Frontier-based exploration using multiple robots,'' in \emph{Proceedings of the second international conference on Autonomous agents}, 1998, pp. 47--53.

\bibitem{burgard2000collaborative}
W.~Burgard, M.~Moors, D.~Fox, R.~Simmons, and S.~Thrun, ``Collaborative multi-robot exploration,'' in \emph{Proceedings 2000 ICRA. Millennium Conference. IEEE International Conference on Robotics and Automation. Symposia Proceedings (Cat. No. 00CH37065)}, vol.~1.\hskip 1em plus 0.5em minus 0.4em\relax IEEE, 2000, pp. 476--481.

\bibitem{pimentel2018information}
J.~M. Pimentel, M.~S. Alvim, M.~F. Campos, and D.~G. Macharet, ``Information-driven rapidly-exploring random tree for efficient environment exploration,'' \emph{Journal of Intelligent \& Robotic Systems}, vol.~91, pp. 313--331, 2018.

\bibitem{zheng_hierarchical_2022}
Z.~Zheng, C.~Cao, and J.~Pan, ``\BIBforeignlanguage{en}{A {Hierarchical} {Approach} for {Mobile} {Robot} {Exploration} in {Pedestrian} {Crowd}},'' \emph{\BIBforeignlanguage{en}{IEEE Robotics and Automation Letters}}, vol.~7, no.~1, pp. 175--182, Jan. 2022.

\bibitem{ibrahim2021enhanced}
M.~F. Ibrahim, A.~B. Huddin, M.~H.~M. Zaman, A.~Hussain, and S.~N. Anual, ``An enhanced frontier strategy with global search target-assignment approach for autonomous robotic area exploration,'' \emph{International Journal of Advanced Technology and Engineering Exploration}, vol.~8, no.~75, p. 283, 2021.

\bibitem{dong_multi-robot_2019}
S.~Dong, K.~Xu, Q.~Zhou, A.~Tagliasacchi, S.~Xin, M.~Niessner, and B.~Chen, ``\BIBforeignlanguage{en}{Multi-{Robot} {Collaborative} {Dense} {Scene} {Reconstruction}},'' \emph{\BIBforeignlanguage{en}{Acm Transactions on Graphics}}, vol.~38, no.~4, p.~84, July 2019.

\bibitem{sun_hierarchical_2023}
X.~Sun, S.~Deng, B.~Tong, S.~Wang, C.~Zhang, and Y.~Jiang, ``Hierarchical framework for mobile robots to effectively and autonomously explore unknown environments,'' \emph{ISA Transactions}, vol. 134, pp. 1--15, Mar. 2023.

\bibitem{kamalova2022occupancy}
A.~Kamalova, S.~G. Lee, and S.~H. Kwon, ``Occupancy reward-driven exploration with deep reinforcement learning for mobile robot system,'' \emph{Applied Sciences}, vol.~12, no.~18, p. 9249, 2022.

\bibitem{alitappeh2022multi}
R.~J. Alitappeh and K.~Jeddisaravi, ``Multi-robot exploration in task allocation problem,'' \emph{Applied Intelligence}, vol.~52, no.~2, pp. 2189--2211, 2022.

\bibitem{perron_or-tools_2023}
L.~Perron and V.~Furnon, ``{OR}-{Tools},'' Mar. 2023.

\end{thebibliography}

\end{document}